\documentclass[acmtog,nonacm]{acmart} 

\usepackage{booktabs} 

\usepackage[ruled]{algorithm2e} 

\SetAlFnt{\small}
\SetAlCapFnt{\small}
\SetAlCapNameFnt{\small}
\SetAlCapHSkip{0pt}

\copyrightyear{2025}
\acmYear{2025}
\setcopyright{acmlicensed}\acmConference[SIGGRAPH Conference Papers '25]{Special Interest Group on Computer Graphics and Interactive Techniques Conference Conference Papers }{August 10--14, 2025}{Vancouver, BC, Canada}
\acmBooktitle{Special Interest Group on Computer Graphics and Interactive Techniques Conference Conference Papers (SIGGRAPH Conference Papers '25), August 10--14, 2025, Vancouver, BC, Canada}
\acmDOI{10.1145/3721238.3730656}
\acmISBN{979-8-4007-1540-2/2025/08}

\newcommand{\pil}{\underline{\pi}}
\newcommand{\pih}{\overline{\pi}}
\newcommand{\E}{\mathbb{E}}

\newcommand{\vect}[1]{\boldsymbol{\mathbf{#1}}}
\newcommand{\w}{\vect{w}}

\usepackage{graphicx}
\usepackage{xcolor}
\usepackage{geometry}
\usepackage{bm}
\usepackage{multirow}
\usepackage{wrapfig}
\usepackage{siunitx}
\usepackage{lipsum,tikz}

\definecolor{upper}{HTML}{6baed6}
\definecolor{lower}{HTML}{4292c6}
\definecolor{feet}{HTML}{2171b5} 
\definecolor{rbs}{HTML}{fd8d3c}
\definecolor{root}{HTML}{41ab5d}
\definecolor{vel}{HTML}{238b45}
\definecolor{smooth}{HTML}{de2d26}

\begin{document}


\title{AMOR: Adaptive Character Control through Multi-Objective Reinforcement Learning}

\author{Lucas N. Alegre}
\authornote{denotes equal contribution.}
\orcid{0000-0001-5465-4390}
\affiliation{
    \institution{Universidade Federal do Rio Grande do Sul}
    \city{Porto Alegre}
    \state{RS}
    \country{Brazil}
}
\affiliation{
    \institution{Disney Research}
    \city{Zürich}
    \country{Switzerland}
}
\email{lnalegre@inf.ufrgs.br}

\author{Agon Serifi}
\authornotemark[1]
\orcid{0000-0003-4439-0023}
\affiliation{
    \institution{Disney Research}
    \city{Zürich}
    \country{Switzerland}
}
\email{agon.serifi@disneyresearch.com}

\author{Ruben Grandia}
\orcid{0000-0002-8971-6843}
\affiliation{
    \institution{Disney Research}
    \city{Zürich}
    \country{Switzerland}
}
\email{ruben.grandia@disneyresearch.com}

\author{David Müller}
\orcid{0009-0001-6591-8803}
\affiliation{
    \institution{Disney Research}
    \city{Zürich}
    \country{Switzerland}
}
\email{david.mueller@disneyresearch.com}

\author{Espen Knoop}
\orcid{0000-0002-7440-5655}
\affiliation{
    \institution{Disney Research}
    \city{Zürich}
    \country{Switzerland}
}
\email{espen.knoop@disneyresearch.com}

\author{Moritz Bächer}
\orcid{0000-0002-1952-1266}
\affiliation{
    \institution{Disney Research}
    \city{Zürich}
    \country{Switzerland}
}
\email{moritz.baecher@disneyresearch.com}

\begin{abstract}
    Reinforcement learning (RL) has significantly advanced the control of physics-based and robotic characters that track kinematic reference motion.
    However, methods typically rely on a weighted sum of conflicting reward functions, requiring extensive tuning to achieve a desired behavior. Due to the computational cost of RL, this iterative process is a tedious, time-intensive task.
    Furthermore, for robotics applications, the weights need to be chosen such that the policy performs well in the real world, despite inevitable sim-to-real gaps.
    To address these challenges, we propose a multi-objective reinforcement learning framework that trains a single policy conditioned on a set of weights, spanning the Pareto front of reward trade-offs.
    Within this framework, weights can be selected and tuned after training, significantly speeding up iteration time.
    We demonstrate how this improved workflow can be used to perform highly dynamic motions with a robot character.
    Moreover, we explore how weight-conditioned policies can be leveraged in hierarchical settings, using a high-level policy to dynamically select weights according to the current task.
    We show that the multi-objective policy encodes a diverse spectrum of behaviors, facilitating efficient adaptation to novel tasks. 
\end{abstract}


\begin{teaserfigure}
    \centering
    \hspace*{-1.172em}\includegraphics[width=1.0352\linewidth]{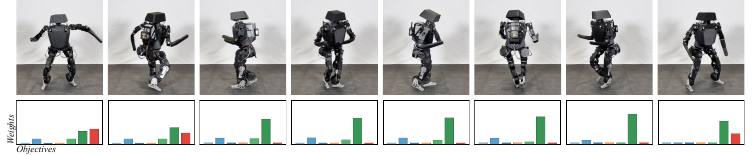}
    \caption{Our method uses multi-objective reinforcement learning to enable on-the-fly tuning of reward weights post-training, which can be used to transfer challenging motions onto physical robots. The bar plots show the tuned weights of the individual reward terms at different points in time: \textcolor{upper}{upper body joint angles}, \textcolor{lower}{lower body joint angles}, \textcolor{feet}{foot joint angles}, \textcolor{rbs}{rigid body poses}, \textcolor{root}{root pose}, \textcolor{vel}{root velocities}, and \textcolor{smooth}{smoothness}.}
    \Description{The teaser figure shows a robot performing a double spin.}
  \label{fig:teaser}
\end{teaserfigure}

\maketitle

\section{Introduction}

Creating motion tracking controllers is a fundamental challenge in physics-based character animation and robotics. The predominant approach trains controllers using reinforcement learning (RL), where weighted sums of carefully-designed reward functions guide the agent towards achieving a desired behavior.

A key challenge is that, due to possibly conflicting rewards (e.g., maximizing accuracy while minimizing energy), choosing the weights is nontrivial. In practice, finding a set of weights that results in the intended behavior often involves a trial-and-error approach. Because standard RL methods require setting these weights prior to training, the required retraining makes this a time-intensive process. Furthermore, in a setting where multiple motions are tracked by a single policy, motions with distinct styles or dynamics may benefit from a different trade-off, which a single fixed set of weights cannot provide.

The requirement for tuning weights is further exacerbated in robotics applications, where a controller is trained in simulation, but expected to perform well in the real world. The sim-to-real gap poses additional, but unknown, requirements on the behavior, which need to be navigated by selecting appropriate weights. For example, smoothness terms typically need to be weighted much higher for satisfactory results on real hardware compared to simulation.

To overcome these limitations, we propose \textbf{A}daptive character control through \textbf{M}ulti-\textbf{O}bjective \textbf{R}einforcement learning (AMOR), a method that leverages multi-objective RL (MORL) to train a control policy conditioned on reward weights. Our method allows users to set the weights after training and directly observe the adapted behavior, without requiring retraining. Any iterative weight tuning is therefore significantly accelerated.

Being able to adjust weights without retraining opens up exciting opportunities. In this paper, we explore two ways to leverage the capabilities of AMOR. First, we manually tune the weights to achieve the sim-to-real transfer of dynamic motions for a robotic character. Second, we explore the use of AMOR in a hierarchical setting, where a high-level policy (HLP) uses its adaptive capabilities to solve a novel task. We observe that reward trade-offs may not only need to vary across but also within skills. To automate this fine-grained selection of reward weights, the HLP dynamically adjusts the reward trade-offs. For training, we rely on a generator-discriminator approach~\cite{ho2016generative, xu2021gan, peng2021amp}.

By enabling adjustments of reward trade-offs without retraining, AMOR paves the way towards adaptive physics-based character control. In summary, our contributions are:
\begin{itemize}
    \item A novel \textit{context-conditioned} MORL problem formulation that enables the extraction of Pareto fronts conditioned on different contexts, with a single policy.
    \item AMOR, a controller conditioned on reward weights and task context, capable of zero-shot adaptation to desired trade-offs among conflicting objectives.
    \item A hierarchical policy that leverages AMOR for fine-grained real-time adjustments of reward weights, offering interpretability of implicit rewards as a byproduct.
\end{itemize}

\section{Related Work}

\paragraph{Physics-based Character Control}

Early work in character control relied on carefully designed cost functions and optimization to synthesize lifelike locomotion and diverse skills~\cite{hodgins1995animating, sok2007simulating, sharon2005synthesis, coros2010generalized, lee2010data,  mordatch2012disovery, mordatch2010robust, borno2013trajectory, hamalainen2015mpc}. 

The growing availability of motion capture data gradually shifted the field towards learning-based methods, where controllers learn from human motion rather than relying on hand-crafted optimization~\cite{liu2010samcon, liu2015sampling, won2017dragon, liu2012terrain}. Especially with the development of efficient RL algorithms~\cite{mnih2015human, schulman2015tppo, mnih2016async, schulman2017ppo, schulman2015high}, neural network controllers gained traction through their ability to learn continuous motor policies from data~\cite{peng2017deeploco, peng2018deepmimic, heess2015continuous, heess2017emergence, merel2018neural, torabi2018behavioral}, although balancing style, realism, and robustness in reward design remains non-trivial.

While some methods have leveraged insights such as symmetry and energy efficiency in locomotion~\cite{yu2018energy}, predominant approaches define objectives that explicitly measure pose and velocity tracking accuracy~\cite{peng2017deeploco, bergamin2019drecon}. Although this has been shown to scale beyond locomotion~\cite{peng2018deepmimic, peng2018sfv}, such methods have often attempted to craft a single reward function for all motions, which has proven challenging and consequently limited their applicability beyond a handful of motions or specific motion styles. 

Various ideas have been explored to scale tracking controllers; combining expert policies~\cite{merel2020catch, won2020scadiver, peng2019mcp, merel2018neural, merel2017learning}, incorporating future frames~\cite{park2019learning, chentanez2018physics}, incrementally increasing motion complexity~\cite{luo2023phc, wang2020unicon}, model-based RL~\cite{fussell2021supertrack, yao2022controlvae}, leveraging latent spaces~\cite{gehring2023leveraging, won2022physics, serifi2024vmp, hasenclever2020comic, zhu2023neural}, or the use of advanced transformer-based architectures~\cite{tessler2024maskedmimic}. Despite this progress, these methods continue to rely on fixed-weight reward design, which remains a critical bottleneck. 

Concurrently, another line of research has attempted to sidestep handcrafted objectives by replacing explicit reward functions with learned implicit rewards~\cite{peng2021amp, xu2021gan}. Instead of explicitly computing deviations, these methods utilize a discriminator score, inspired by adversarial learning~\cite{ho2016generative, goodfellow2014generative, torabi2018generative}, to differentiate between the reference and the controller-produced simulated motion. While they automate aspects of reward construction, they introduce training instabilities like mode collapse and require significant compute. Moreover, they lose the interpretability of explicit rewards and are usually used to design controllers that aim to embed a repertoire of diverse skills into the policy rather than tracking a specific reference motion well. Some limitations were improved by follow-up work~\cite{tessler2023calm, dou2023case, tang2024humanmimic}, but challenges remain.

Recently, efforts have been made to bring physics-based controllers to robots~\cite{serifi2024vmp, cheng2024expressive, he2024hover}, where reward design has proven even more challenging~\cite{ibarz2021how, ha2024learning, gu2025humanoid}. Potential sim-to-real gaps can result in unexpected trade-offs, requiring an adjustment of reward functions that, in turn, necessitates a retraining of the controller from scratch. 

\paragraph{Multi-Objective Reinforcement Learning (MORL)} 

Multi-objective optimization and Pareto front extraction have seen use in interactive design exploration in graphics~\cite{schulz2018interactive}. In our work, we highlight applications in RL and physics-based character control instead.

MORL~\cite{hayes2022practical} emerged as an extension of traditional RL~\cite{sutton2018rl} to address problems involving multiple conflicting objectives. Unlike standard RL, where a single policy is trained, agents are tasked to learn a \emph{set} of policies, each specializing to a different prioritization between the reward functions. Early MORL work targeted the development of the underlying theory and the training of small sets of Pareto-optimal policies in synthetic environments~\cite{roijers2013survey,vanmoffaert&nowe2014}.

Recent MORL work can be divided into techniques that explicitly maintain a population of policies~\cite{xu2020prediction,felten+2024}, and methods that learn a single preference-conditioned model that encodes multiple policies~\cite{yang2019generalized,alegre2023sample}. 
Our work falls into the second category, which scales better with the number of policies. 

A few works have considered constrained MORL approaches for physics-based character motion~\cite{wang2020unicon,kim2024stagewise}. In Wang et al.~\shortcite{wang2020unicon}, tasks are formulated using multi-objective reward functions. However, fixed weights and cost thresholds are employed, limiting the versatility of the resulting controllers.
More recent work models tasks by dividing them into pre-specified stages, defined by different rewards and cost functions~\cite{kim2024stagewise}. However, this method requires an expert to manually specify the cost thresholds and reward weights for each pre-specified stage of a task. To employ multiple critics, Xu et al.~\shortcite{xu2023composite} use a MORL formulation where different groups of body parts are treated as independent tasks with their own value function. However, the authors only learn a single policy on the Pareto front, assigning fixed weights for all body groups. Our method, in contrast to the aforementioned works, learns trade-offs that cover the entire space of preferences. 

Another setting related to MORL is the multi-task setting within the successor features (SFs) framework ~\cite{dayan1993successor,barreto2017successor,barreto2019option}. In this line of work, the reward function is assumed to be a linear combination of a set of (learned) \textit{features}. \citet{alegre2022} showed connections between the MORL and SFs settings, demonstrating that ideas from both fields could be combined.
While SF-based methods focus on the rapid adaptation to novel tasks, the goal in MORL---and in this paper---is to learn a set of Pareto-optimal solutions that capture all trade-offs between conflicting objectives.

\section{AMOR Overview}

\begin{figure}
    \centering
    \includegraphics[width=\linewidth]{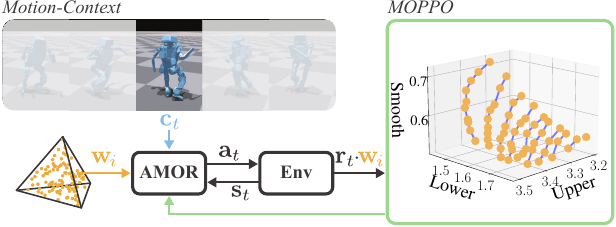}
    \caption{\textbf{AMOR Overview.} AMOR optimizes for multiple objectives conditioned on state, motion-context, and reward weights, where reward weights are sampled from a multi-dimensional simplex. The environment provides a vector of rewards, which are then used by a multi-objective PPO algorithm together with the weights to update the policy.}
    \label{fig:overview}
    \Description{overview}
\end{figure}

Fig.~\ref{fig:overview} shows the structure of AMOR. At its core, it is an RL-based controller that tracks a kinematic reference motion, similarly to related work \cite{serifi2024vmp}. The policy is trained to output actions $\mathbf{a}_t$ that maximize the reward of a simulated character, given the current state of the character $\mathbf{s}_t$, and a context vector $\mathbf{c}_t$ encoding task-relevant information. Specifically, in our context, $\mathbf{c}_t$ includes a time-varying kinematic reference, and also a latent-space encoding of a motion window that captures past and future targets.

The problem specification also includes a set of reward terms $\mathbf{r}_t$ that capture the policy performance. Particularly, we consider a set of 7 rewards, including joint-space and task-space tracking, velocity tracking, and smoothness. Instead of prescribing a fixed set of reward weights, and giving the policy a scalar reward, we maintain a vector of rewards and condition the policy on the reward weights, so the policy becomes $\pil(\mathbf{a}_t|\mathbf{s}_t,\mathbf{c}_t,\vect{w})$. 

For each training episode and environment, we sample a set of reward weights from a multi-dimensional simplex. Policy updates are then computed using a multi-objective extension of the PPO algorithm, which is also given this same weight vector. Once converged, this results in a policy which can track arbitrary reference motions under arbitrary weightings of different reward terms, allowing for on-the-fly weight tuning post-training by either a user or an algorithm.

\section{Multi-Objective Reinforcement Learning}

Before discussing AMOR and its HLP extension in more detail, we provide background on multi-objective reinforcement and introduce our algorithmic extensions. 

\subsection{Background}

In standard RL~\cite{sutton2018rl}, an agent interacts with its environment by selecting actions $\mathbf{a}_t$ based on the current state $\mathbf{s}_t$ following a policy $\pi(\mathbf{a}_t|\mathbf{s}_t)$. This action leads to a transition to a new state $\mathbf{s}_{t+1} \sim p(\cdot|\mathbf{s}_t,\mathbf{a}_t)$ and yields a scalar reward $r_t=r(\mathbf{s}_t,\mathbf{a}_t,\mathbf{s}_{t+1})$. The agent's goal is to maximize the expected discounted return, 
\begin{equation}
J(\pi) = \E_{\mathbf{d}_0} \left[ V^\pi(\mathbf{s}_0) \right]= \E_\pi \left[ \sum_{t\geq0} \gamma^t r_t \mid \mathbf{s}_0 \sim \mathbf{d}_0\right],     
\end{equation}
where $\gamma \in [0,1)$ is the discount factor, $V^\pi(\mathbf{s})$ is the value function, and $\mathbf{d}_0$ an initial state distribution. 

MORL~\cite{hayes2022practical} extends this framework by providing the agent with a vector-valued reward $\vect{r}_t(\mathbf{s}_t, \mathbf{a}_t, \mathbf{s}_{t+1}) \in \mathbb{R}^m$, where each element represents a distinct (and potentially conflicting) objective. Instead of a scalar return, each policy is associated with an expected vector return, $\vect{J}(\pi) = \E_\pi \left[ \sum_{t\geq0} \gamma^t \vect{r}_t  \mid \mathbf{s}_0 \sim \mathbf{d}_0 \right]$. In contrast to standard RL, no single optimal solution exists; instead, the goal of an agent is to identify a \textit{Pareto front}, $\mathcal{F}$. 
We say a point $\vect{J}(\pi)$ is \textit{Pareto non-dominated} if and only if there does not exist another point $\vect{J}(\pi')$ such that $J_i(\pi') \geq J_i(\pi), \forall i$ and $J_i(\pi') > J_i(\pi)$ for at least one $i\in\{1,...,m\}$.
In the context of continuous robotic control tasks~\cite{xu2020prediction,felten2023toolkit}, the Pareto Front is typically convex\footnote{In the MORL literature, the Pareto front under linear preferences is also known as the \textit{convex coverage set} (CCS)\cite{roijers2013survey}.} and can be defined in terms of a linear dominance relation
\begin{equation}
\label{eq:pareto_front}
    \mathcal{F} = \left\{ \vect{J}(\pi) \mid \exists \w \ \text{s.t.} \  \vect{J}(\pi) \cdot \w \geq \vect{J}(\pi') \cdot \w, \forall \pi' \right\},
\end{equation}
where the elements $w_i$ in the reward weight vector $\w\in\Delta^m$ form a convex combination, satisfying the requirements $\sum_{i=1}^{m} w_i = 1$ and $w_i \geq 0$.
Intuitively, each $\w$ induces a different scalar reward $r_t = \vect{r}(\mathbf{s}_t, \mathbf{a}_t,\mathbf{s}_{t+1}) \cdot \w$. Due to the linearity of the expectation and sum operations, it follows that
\begin{equation}
J(\pi) = \E_\pi \left[ \sum_{t\geq0} \gamma^t \vect{r}_t  \cdot \w \right] = \E_\pi \left[ \sum_{t\geq0} \gamma^t \vect{r}_t   \right] \cdot \w = \vect{J}(\pi) \cdot \w,
\end{equation}
with the optimal solution $J^{*} = \max_\pi \vect{J}(\pi) \cdot \w$. The corresponding policy $\pi^{*}$ is therefore optimal for any trade-off $\mathbf{w}$ between the $m$ objectives.

\subsection{Algorithmic Extensions}

We extend the standard MORL framework by conditioning policies, $\pi(\mathbf{a}_t|\mathbf{s}_t, \mathbf{c}_t, \mathbf{w})$, and rewards, $\vect{r}_t(\mathbf{s}_t, \mathbf{a}_t, \mathbf{s}_{t+1}, \mathbf{c}_t)$,  on an additional context vector $\mathbf{c}_t$ that encodes task-relevant information, making the Pareto front also a function of the context. Given a context, each $\mathbf{w}\in\Delta^m$ then induces a different Pareto-optimal policy within the Pareto front.

To train $\pi$, we introduce a multi-objective extension of the Proximal Policy Optimization (PPO) algorithm~\cite{schulman2017ppo}, which we refer to as MOPPO. Instead of learning a scalar value function $V^\pi(\mathbf{s}, \mathbf{c}) = \E_{\pi} \left[ \sum_{t\geq0} \gamma^t r_t \mid \mathbf{s}_0 = \mathbf{s}, \mathbf{c}_0 = \mathbf{c} \right]$, as done in traditional goal-conditioned RL, MOPPO learns a vector-valued function conditioned on the reward weights $\w$,
\begin{equation}
    \vect{V}^\pi(\mathbf{s},\mathbf{c},\w) = \E_{\pi} \left[ \sum_{t\geq0} \gamma^t \vect{r}_t \mid \mathbf{s}_0 = \mathbf{s}, \mathbf{c}_0=\mathbf{c}\right],
\end{equation}
optimizing multiple objectives simultaneously. The multi-objective policy gradient, $\nabla_\pi \left[\vect{J}(\pi) \cdot \w \right]$, used to update the policy,  
\begin{equation}
     \E_{d^\pi} \left[ \sum_{t\geq0}  ( \vect{A}^\pi(\mathbf{s}_t,\mathbf{c}_t,\mathbf{a}_t) \cdot \w ) \ \nabla_\pi \log \pi(\mathbf{a}_t|\mathbf{s}_t,\mathbf{c}_t,\w) \right],
     \label{eq:mo-objective}
\end{equation}
relies on a vector-valued advantage function $\vect{A}^\pi(\mathbf{s}_t,\mathbf{c}_t,\mathbf{a}_t)$. Here, $d^\pi$ represents the discounted stationary distribution of states induced by $\pi$ and the environment dynamics, $\mathbf{s}_{t+1},\mathbf{c}_{t+1}\sim p(\cdot|\mathbf{s}_t,\mathbf{c}_t,\mathbf{a}_t)$. As is common practice in PPO implementations~\cite{andrychowicz2021what}, we employ generalized advantage estimation (GAE) and normalize the scalarized advantage, $\vect{A}^\pi\cdot\w$, with its mean and standard deviation over each mini-batch. Moreover, we employ the standard PPO clipped loss function, but constructed based on Eq.~\eqref{eq:mo-objective}.

To ensure coverage of the space of possible reward weights and the Pareto front, we assign a different randomly-sampled weight vector $\w \sim \Delta^m$ to every episode. As a result, the agent's replay buffer stores experience transitions of the form $(\mathbf{s}_t,\mathbf{c}_t,\mathbf{a}_t,\vect{r}_t,\mathbf{s}_{t+1},\mathbf{c}_{t+1},\w)$.

To sample weight vectors uniformly from the 
$(m{-}1)$-dimensional simplex $\Delta^m$, we draw from a Dirichlet distribution with parameter $\alpha=1$.

\section{Training of AMOR Policy}

To instantiate AMOR for character control tasks, we train a multi-objective policy to track reference motions under multiple, potentially conflicting objectives. This section details how we formulate the tracking problem, construct the motion context, define the multi-objective reward, and incorporate the reward weights into our policy.

\paragraph{Motion Tracking Problem}

We consider the standard problem of tracking kinematic motion using a physically simulated character. As in VMP~\cite{serifi2024vmp}, we assume access to a dataset $\mathcal{D}$ of motion clips consisting of a finite sequence of motion frames, $\mathbf{m}_t = (h_t,\bm{\theta}_t,\mathbf{v}_t,\mathbf{q}_t,\dot{\mathbf{q}}_t,\mathbf{p}_t, \dot{\mathbf{p}}_t)$. Here, $h_t$ is the height of the character's root relative to the ground, $\bm{\theta}_t$ is the orientation of the root in a 6D representation~\cite{zhou2019continuity, xiang2020revisiting}, and $\mathbf{v}_t$ is a 6D vector representing the root's linear and angular velocities. Moreover, $\mathbf{q}_t$ is the angular position of the character's joints and $\dot{\mathbf{q}}_t$ their angular velocities. Finally, $\mathbf{p}_t$ is a 9D vector that encodes the poses of hands and feet relative to the root (3D position, 6D orientation) and $\dot{\mathbf{p}}_t$ encodes the corresponding linear velocities. To make the motion invariant to the global pose, we normalize frames $\mathbf{m}_t$ by expressing all orientations and velocities with respect to the local heading frame of the root $\bm{\theta}_t$.

Our goal is to choose actions $\mathbf{a}_t$ to match these reference poses as closely as possible, subject to realistic physics constraints. However, perfect tracking of reference motion in the dataset is generally infeasible. Largely due to an ill-posed retargeting from human mocap to our characters, there are artifacts such as imbalance, inter-penetrations with the ground, or foot sliding that are clearly visible to the naked eye. Thus, certain aspects of a kinematic reference may need to be prioritized to maintain dynamic feasibility, creating conflicting objectives.

\paragraph{Motion Context} To condition the policy on the task, we define the \textit{motion context} $\mathbf{c_t} = (\mathbf{m}_t, \mathbf{z}_t)$, where we add a latent vector $\mathbf{z}_t$ that represents a compressed motion window of frames $\mathbf{M}_t = \{\mathbf{m}_{t-W},...,\mathbf{m}_{t+W}\}$, centered at $t$ and of size $2W+1$, alongside the current motion frame. For training of this latent representation, we follow  VMP~\cite{serifi2024vmp} and train a Variation Autoencoder (VAE) that maps motion windows to latent codes $\mathbf{z}_t = e(\mathbf{M}_t)$. This latent representation captures local motion patterns around the frame $t$. We refer to~\cite{serifi2024vmp} for details on the loss function, the training of the VAE, and the normalization of motion windows.

\paragraph{Multi-Objective Tracking Reward} We use a multi-objective reward vector $\vect{r} \in \mathbb{R}^m$ with $m=7$ motion tracking objectives
\begin{equation}
\label{eq:mo-tracking-reward}
    \vect{r}(\mathbf{s}_t,\mathbf{a}_t,\mathbf{s}_{t+1},\mathbf{c}_t) = [r^{\mathrm{up}}_t, r^{\mathrm{lo}}_t, r^{\mathrm{feet}}_t, r^{\mathrm{rbs}}_t, r^{\mathrm{root}}_t, r^{\mathrm{vel}}_t, r^{\mathrm{smooth}}_t]^\top.
\end{equation}
The reward $r^{\mathrm{up}}$ tracks the character's upper joint positions and its height, $r^{\mathrm{lo}}$ the lower joint positions, $r^{\mathrm{feet}}$ the positions of the ankle joints, and $r^{\mathrm{rbs}}$ the position and orientation of the end-effectors. Reward $r^{\mathrm{root}}$ tracks the orientation of the root, $r^{\mathrm{vel}}$ the linear and angular velocities of the joints and root, and $r^{\mathrm{smooth}}$ penalizes high action rates and torques to mitigate vibrations and smoothen the motion.

Since the rewards may vary significantly in magnitude, directly summing them up using linear weights could lead to poor problem conditioning. We therefore assign a prior scaling to each reward term, informed by \cite{serifi2024vmp}. 

The individual reward terms and their prior scaling factors are detailed in Tab.~\ref{tab:objectives}.
All objectives include a constant survival bonus, $c^{\mathrm{alive}}$, that rewards the agent for not reaching a terminal state (e.g., falling to the ground). This is important for training to prevent the policy from terminating as quickly as possible to avoid any negative accumulation of reward.

\begin{table}
\centering
\caption{\textbf{Multi-Objective Components.} The individual terms and scales of each of the seven objectives. Scaling values are reported separately for the humanoid and the robot. $\mathbf{q}^{\mathrm{up}|\mathrm{lo}|\mathrm{feet}}$ denote the DoFs for the upper body, lower body, and feet, respectively. $\mathbf{p}^{\text{pos|rot}}$ represent the positional and rotational parts of the rigid body poses, respectively. $\mathcal{R}: \mathbb{R}^n \rightarrow SO(3) \subset \mathbb{R}^{3 \times 3}$ represents the transformation that maps quaternions or 6D rotation representations to their corresponding rotation matrix in $SO(3)$. $\mathbf{v}^{\mathrm{lin}|\mathrm{ang}}$ respectively denote the linear and angular velocity of the root. $\bm{\tau}$ denotes joint torques and $\ddot{\mathbf{q}}$ joint accelerations. Target values are denoted by $(\hat{\cdot})$.}
\begin{footnotesize}
\begin{tabular}{l l rr}
\toprule
 & & \multicolumn{2}{c}{\hspace{2.2em}\textbf{Scale}}\\
\textbf{Objective} & \textbf{Term(s)} & \textbf{Humanoid} & \textbf{Robot} \\
\midrule
\multirow{1}{*}{$r^{\mathrm{up}}$} &  $\|\mathbf{q}^{\mathrm{up}}-\hat{\mathbf{q}}^{\mathrm{up}}\|_2^2$   & 1.0 &  7.0\\
\midrule
\multirow{1}{*}{$r^{\mathrm{lo}}$}   &  $\|\mathbf{q}^{\mathrm{lo}}-\hat{\mathbf{q}}^{\mathrm{lo}}\|_2^2$   & 1.0 & 7.0 \\
\midrule
\multirow{1}{*}{$r^{\mathrm{feet}}$}   & $\|\mathbf{q}^\mathrm{feet}-\hat{\mathbf{q}}^\mathrm{feet}\|_2^2$    & 1.0 & 7.0 \\
\midrule
\multirow{3}{*}{$r^{\mathrm{rbs}}$}   &  $\|\mathbf{p}^\text{pos}-\hat{\mathbf{p}}^\text{pos}\|_2^2$  & 1.0 & 1.0 \\
 &  $\|\mathcal{R}(\mathbf{p}^\text{rot})-\mathcal{R}(\hat{\mathbf{p}}^\text{rot})\|_2^2$  & 1.0 & 1.0 \\
  &  $\|\dot{\mathbf{p}}-\hat{\dot{\mathbf{p}}}\|_2^2$  & 1.0 & 1.0 \\
\midrule
\multirow{1}{*}{$r^{\mathrm{root}}$}   & $\|\mathcal{R}(\mathbf{\bm{\theta}})-\mathcal{R}(\hat{\mathbf{\bm{\theta}}}) \|_2^2$   & 1.0 & 1.0\\
\midrule
\multirow{2}{*}{$r^{\mathrm{vel}}$}   & $\|\mathbf{v}^\mathrm{lin}-\hat{\mathbf{v}}^\mathrm{lin}\|_2^2$   & 1.0 & 2.0 \\
&  $\|\mathbf{v}^\mathrm{ang}-\hat{\mathbf{v}}^\mathrm{ang}\|_2^2$  & 1.0 & 2.0 \\
\midrule
\multirow{4}{*}{$r^{\mathrm{smooth}}$}   & $-\|\bm{\tau}\|_2^2$      & $ 1.0 \cdot 10^{-5}$ & $ 1.0 \cdot 10^{-4}$ \\
  &   $-\|\mathbf{a}_t-\mathbf{a}_{t-1}\|_2^2$                       & $ 1.0 \cdot 10^{-5}$  & $ 1.5$ \\
  &   $-\|\mathbf{a}_t-2\mathbf{a}_{t-1}+\mathbf{a}_{t-2}\|_2^2$  \hspace{0.5cm}   & $ 1.0 \cdot 10^{-5}$ & $ 0.45$ \\
  &   $-\|\ddot{\mathbf{q}}\|_2^2$                                   & $ 1.0 \cdot 10^{-6}$ & $ 2.5 \cdot 10^{-6}$ \\
\bottomrule
\end{tabular}
\end{footnotesize}
\label{tab:objectives}
\end{table}

\paragraph{Weight Conditioning} 

In contrast to traditional motion tracking policies, AMOR is additionally conditioned on reward weight vectors $\w$. For each environment and each episode, we sample weights in the simplex $\Delta^m$ and keep them fixed while the motion context moves forward in time but also jumps to new motions. After collecting enough samples, gradient steps are performed to update the policy according to the MOPPO objective (Eq.~\ref{eq:mo-objective}).

\section{Hierarchical Weight Adjustment}
\label{sec:hierarchical_weight_adjustement}

\begin{figure}[!t]
    \centering
    \includegraphics[width=\linewidth]{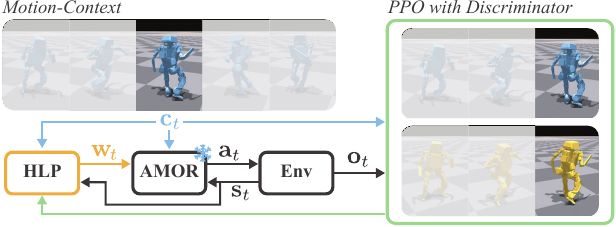}
    \caption{\textbf{High-Level Policy Overview.} We learn a high-level policy (HLP) that generates reward weights for a pretrained AMOR based on the current motion context. In this stage, a discriminator is trained to distinguish between reference and simulated motions, with its output serving as an implicit reward for the HLP.}
    \label{fig:hlp_overview}
    \Description{HLP overview}
\end{figure}

AMOR's policy, described in the previous section, results in different behavior, dependent on the input weight vector $\w$. This allows the user to flexibly select the desired trade-offs in a \textit{zero-shot} manner, without having to train different agents from scratch. In this section, we alternatively propose a \textit{high-level policy} (HLP), $\pih(\vect{w}_t|\mathbf{s}_t, \mathbf{c}_t)$, which dynamically selects weights $\vect{w}_t$ during execution, enabling the agent to adapt its behavior to the current context and dynamic state based on a different, high-level reward. This hierarchical approach uses AMOR with frozen parameters as its low-level policy and is illustrated in Fig.~\ref{fig:hlp_overview}. We train the HLP using standard PPO, with the addition of a softmax activation function in the actor network's final layer to ensure it outputs reward weights in the simplex, i.e., $\w_t \in \Delta^m$.
We note that training the HLP is significantly faster than training AMOR, as is typically the case in hierarchical methods~\cite{barreto2019option}.

\paragraph{Implicit Reward} The HLP is trained on a reward function, which is not required to be part of the original reward terms. To demonstrate this, we use an implicit reward, as proposed by recent work~\cite{peng2021amp, tessler2023calm}: a discriminator tries to distinguish simulated motions from the kinematic reference motions, and a reward is computed based on accuracy. The hierarchical weight adjustment method allows the expansion of AMOR to optimize for such new rewards. 

We define $\mathbf{O}_t = \{\mathbf{o}_{t-V},...,\mathbf{o}_{t}\}$ as a window of size $V$, containing past observations $\mathbf{o}_t = (\bm{\theta}_t, \mathbf{v}_t, \mathbf{q}_t)$. Let $D(\mathbf{O}_t|\mathbf{z}_t)$ be a discriminator whose goal is to distinguish between dataset transitions $\hat{\mathbf{O}}_t \sim d^{\mathbf{M}}(\hat{\mathbf{O}}_t,\mathbf{z}_t)$ and transitions resulting from following a given policy, $\mathbf{O}_t \sim d^{\pi}(\mathbf{O}_t,\mathbf{z})$, where $d^{\mathbf{M}}(\hat{\mathbf{O}}_t,\mathbf{z}_t)$ and $d^\pi(\mathbf{O}_t,\mathbf{z}_t)$ are state transition distributions of the reference motion and the policy, respectively. The discriminator-based reward function used to train the high-level policy is then
\begin{equation}
\label{eq:discriminator_reward}
    r^D(\mathbf{O}_t,\mathbf{z}_t) = - \log(1 - D(\mathbf{O}_t | \mathbf{z}_t)).
\end{equation}
Intuitively, this reward function rewards the agent for performing transitions that appear indistinguishable from the real motion transitions in the dataset $\mathcal{D}$.
By maximizing this reward function, $\pih$ allows the agent to select the reward weights $\w_t$ that lead to more realistic motion transitions for each $\mathbf{s}_t$ and motion context $\mathbf{c}_t$.

We follow \citet{tessler2023calm} and also condition the discriminator on the latent motion vector $\mathbf{z}_t$ to prevent mode collapse.
The discriminator is trained via the loss function
\begin{equation} \label{eq:discriminator_loss}
L^D = -\E_{\mathbf{M}_t\in\mathcal{D}} \left[  L^{\mathbf{M}} + L^{\pi} + c^{\text{gp}}L^{\text{gp}} \mid \mathbf{z}_t = e(\mathbf{M}_t) \right],
\end{equation}
with terms
\begin{equation} \nonumber
\begin{split}
L^{\mathbf{M}} &= \E_{d^{\mathbf{M}}(\hat{\mathbf{O}}_t,\mathbf{z}_t)} \log D(\hat{\mathbf{O}}_t|\mathbf{z}_t) \\
L^{\pi} &= \E_{d^{\pi}(\mathbf{O}_t,\mathbf{z}_t)} \log(1 - D(\mathbf{O}_t|\mathbf{z}_t)) \\
    L^{\text{gp}} &= \E_{d^{\mathbf{M}}(\hat{\mathbf{O}}_t,\mathbf{z}_t)} \|\nabla_{\phi}D(\phi)\mid_{\phi=(\hat{\mathbf{O}}_t,\mathbf{z_t})}\|^2 ,
\end{split}
\end{equation}
which has shown to minimize the Jensen-Shannon divergence between $d^{\mathbf{M}}(\hat{\mathbf{O}}_t,\mathbf{z}_t)$ and $d^\pi(\mathbf{O}_t,\mathbf{z}_t)$~\cite{nowozin2016fgan}. $L^{\text{gp}}$ is a gradient penalty, scaled by coefficient $c^{\text{gp}}$, used to penalize nonzero gradients on samples from the dataset. This has been shown to improve training stability~\cite{mescheder2018training}.

Note that attempts to directly use the discriminator-based reward in the low-level policy led to mode collapse; the presented two-stage approach seems better suited for navigating the complex landscapes created by such higher-level rewards.

\paragraph{Reward Interpretability} After training the HLP on the high-level reward, we can inspect the weights it selects. This shows which combination of low-level reward terms correspond to a behavior that maximizes the implicit reward. The combination of the HLP and AMOR thus allows us to interpret what the discriminator is looking for at a given state and context. For additional artistic control, a user could also edit the weights returned by the HLP.

\section{Evaluation and Results}

In this section, we evaluate our method's capabilities for the problem of motion tracking of physically-based characters and robots. In particular, we aim to validate that \textit{(i)} AMOR is able to approximate the Pareto front of tracking behavior for different motions with a single policy, allowing for behavior tuning without retraining; \textit{(ii)} the relative importance between tracking objectives plays a significant role in the resulting tracking behavior;  and \textit{(iii)} by employing our high-level policy, $\pih(\w|\mathbf{s_t},\mathbf{c_t})$, we can dynamically prioritize different objectives resulting in more flexible and robust tracking behavior.

\subsection{Experimental Setting}

\paragraph{Characters} We perform our experiments on a standard humanoid with 36 degrees of freedom (DoFs) and a bipedal robot with 20 DoFs. The characters are torque-controlled using a proportional-derivative (PD) controller with realistic actuator models for the robot~\cite{grandia2024design} and virtual actuators for the humanoid~\cite{serifi2024vmp}.

\paragraph{Dataset} The dataset consists of motion capture data from a simple mocap setup (CMU~\shortcite{cmu}, 1870 clips, \SI{8.5}{\hour}) as well as a smaller high-quality dataset of processed motions (Reallusion~\shortcite{reallusion}, 214 clips, \SI{0.5}{\hour}).

\paragraph{RL} We employ multi-layer perceptron (MLP) neural networks with ELU activations~\cite{clevert2016fast} to model the policies and value functions. We use 4 layers with 1024 units to model AMOR $\pil$ and the critic, and a 3-layer model for the high-level policy $\pih$. Both policies operate at \SI{50}{\hertz}.
We normalize the observations using a running mean, as typically done when using PPO~\cite{andrychowicz2021what}.
Our simulations are conducted using the GPU-accelerated Isaac Gym~\cite{makoviychuk2021isaacgym} simulator, running 8192 environment instances in parallel at \SI{250}{\hertz} on a single RTX 4090 GPU. We train AMOR for 300k iterations (approximately 5 days) for each character.

\subsection{AMOR}

\paragraph{Pareto Front}

\begin{figure}
    \centering
    \includegraphics[width=0.92\linewidth]{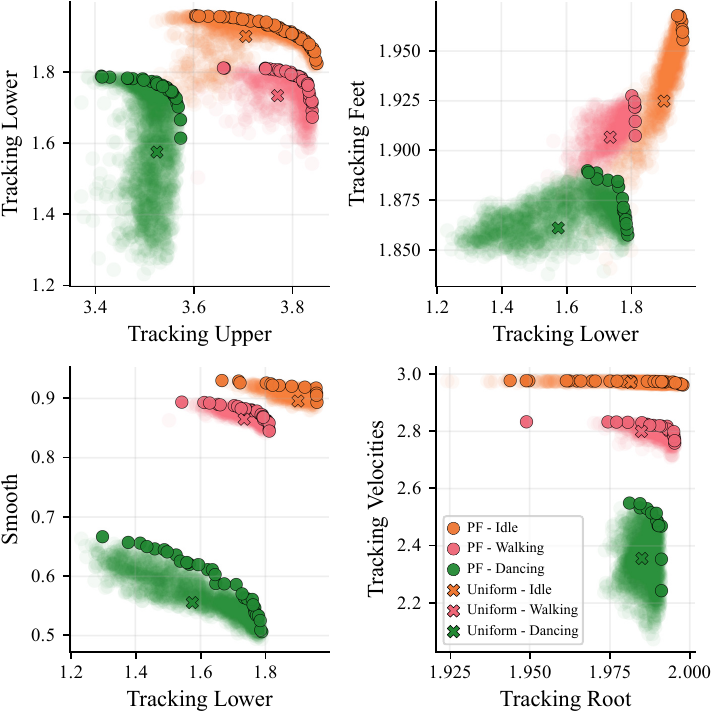}
    \caption{\textbf{Pareto Fronts (PFs)}. Visualization of selected PFs generated by tracking three distinct motion types—\textcolor[rgb]{0.933, 0.466, 0.2}{Idle}, \textcolor[rgb]{0.933, 0.400, 0.467}{Walking}, and \textcolor[rgb]{0.133, 0.533, 0.200}{Dancing}—using the humanoid controlled by AMOR's policy $\pil$. \textbf{x}-markers indicate performance under equal weight configuration, corresponding to a fixed-reward policy.}
    \label{fig:huamnoid-pfs}
    \Description{Humanoid Pareto fronts of different motions.}
\end{figure}

\begin{figure}
{
\includegraphics[width=\linewidth]{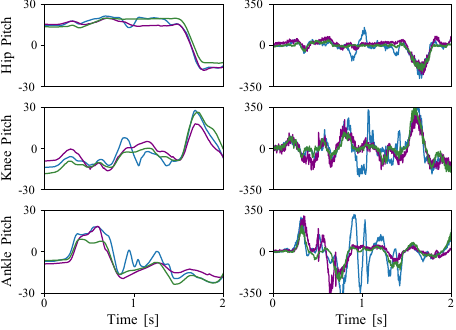}}
\caption{\textbf{Weight Tuning}. The left column shows the measured DoF $[^\circ]$, and the right column shows the measured DoF velocities $[^\circ/\mathrm{s}]$ for the three pitch joints of the robot's right leg during the dance motion in the supplementary video. The 
\textcolor[rgb]{0.502, 0.0, 0.502}{purple curve} shows the measurements of the simulated robot and the \textcolor[rgb]{0.1216, 0.4667, 0.7059}{blue curve} of the real robot with uniform weight distribution. Through tuning the weights and increasing the smoothness objective, the sim-to-real gap and undesired jitter can be reduced, as shown with the \textcolor[rgb]{0.2225, 0.5275, 0.2225}{green curve}. See also the supporting video.}
\label{fig:robot_weight_tuning_joints}
\Description{Humanoid dancing through weights}
\end{figure}

First, we show a visualization of the Pareto front identified by AMOR's policy, when given different motions to track. In particular, we evaluate $\tilde{\mathcal{F}} = \{\vect{J}(\pil(\cdot,\w)) \mid \w \sim \Delta^m\}$ by sampling 8192 weight vectors from $\Delta^m$ and averaging over 15 episodic returns.
Fig.~\ref{fig:huamnoid-pfs} displays the Pareto fronts obtained by tracking three different motions (Idle, Walking, Dancing) on the humanoid by following $\pil$.
Because each point $\vect{J}(\pi)$ on the Pareto front is a 7-dimensional vector, we depict pairwise comparisons between the unscaled cumulative reward corresponding to each objective. 
All of the 8192 points are Pareto non-dominated w.r.t. all objectives, and the points with a black border are points that are additionally Pareto non-dominated w.r.t. the two objectives in the corresponding figure panel. We also depict with crosses the mean return obtained by following $\pil$ with equal reward weights, $w_i= 1/m$. Although the reward terms share the overall goal of achieving better motion tracking, they are inherently conflicting. For example, for the dancing motion in Fig.~\ref{fig:huamnoid-pfs}, we observe that prioritizing smoothness conflicts with lower-body tracking. Precisely tracking more dynamic reference motions, which are generally less feasible, introduces jitter. This trade-off is also visible on the physical system, see Fig.~\ref{fig:robot_weight_tuning_joints}. Even seemingly unrelated objectives, such as upper- and lower-body tracking, can conflict because the coordination between the body parts is not physically accurate.
We note that different motions induce different Pareto fronts with varying degrees of conflict between objectives. This suggests that there is indeed value in not relying on fixed weights when tracking multiple motions. 

\paragraph{Training Comparison}

\begin{wrapfigure}{r}{0.20\textwidth}
    \centering
    \includegraphics[width=1.0\linewidth]{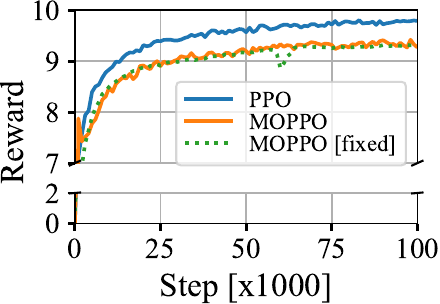}
\end{wrapfigure}

We compare the total reward obtained during training between AMOR trained with MOPPO and a policy trained with PPO, using the fixed average weights, $w_i= 1/m$. For MOPPO, we evaluate the reward using both randomly sampled weights and the fixed weights used in PPO training. In all cases, we report the sum of equally weighted reward terms. As shown in the inset figure, the training follows a similar trend, with PPO converging faster. This faster convergence is expected, as PPO optimizes for a \textit{single} fixed reward weight combination, while MOPPO must dynamically adapt its behavior to varying weights covering the entire simplex $\Delta^m$, making it a harder learning task. We hypothesize that the gap in total reward between PPO and MOPPO would further reduce after both algorithms fully converge. Notably, MOPPO's reward evaluated for the average weights is similar to the average reward across the randomized weights.

\paragraph{Behavior Adaptation}

\begin{figure}[b]
{
\includegraphics[width=\linewidth]{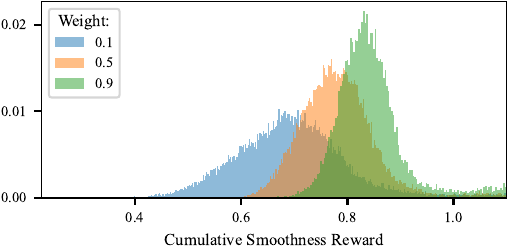}}
\caption{\textbf{Prioritizing Objectives}. Distributions of the unweighted cumulative smoothness reward for three distinct weight values, each evaluated on 32768 episodes.}
\label{fig:cum_sum}
\Description{Prioritizing Objectives}
\end{figure}

To validate that changes in weights indeed lead to a change in behavior, we manually change the weights for the humanoid while performing a dancing motion. The results are best seen in the supplementary video. Fig.~\ref{fig:humanoid_dancing_through_weights} highlights keyframes of the experiment. It can be seen that with the increased smoothness term, a smoother motion is obtained at the expense of tracking performance, as would be expected. We further evaluate this observation on the full dataset; Fig.~\ref{fig:cum_sum} shows the distribution of the unweighted cumulative reward for smoothness when increasing the corresponding reward weight. The change in the distribution shows that the policy can successfully adapt its behavior to prioritize smoothness.

\begin{figure}[b]
{
\setlength{\tabcolsep}{0.8pt}
\begin{tabular}{ccccc}
 & 
\includegraphics[width=0.07\textwidth]{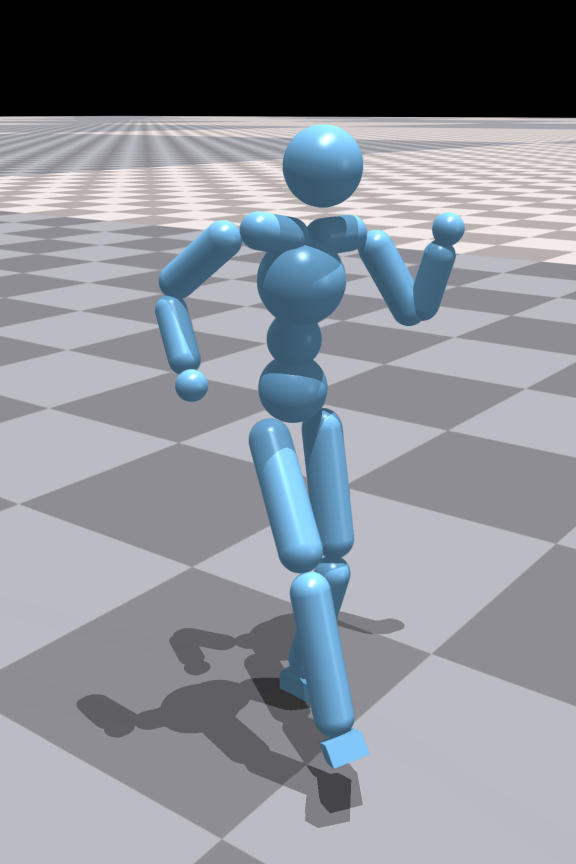} & 
\includegraphics[width=0.07\textwidth]{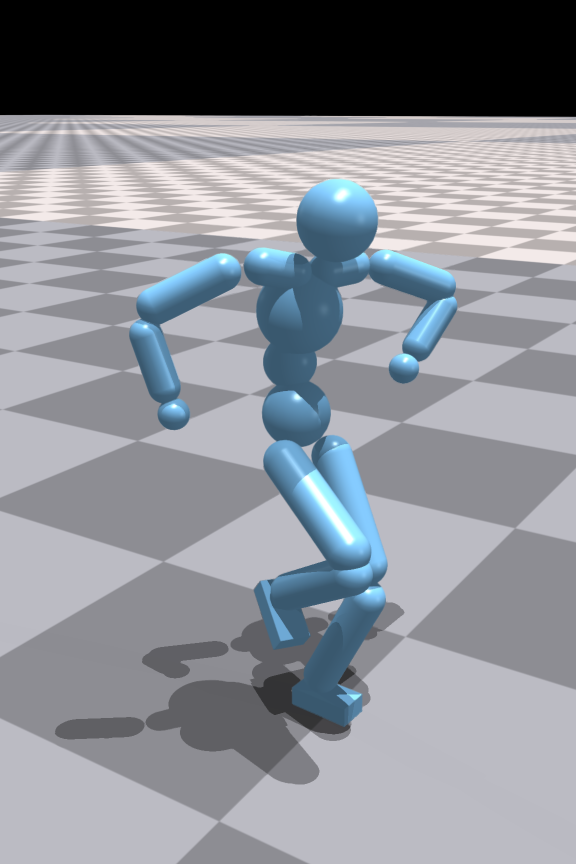} & 
\includegraphics[width=0.07\textwidth]{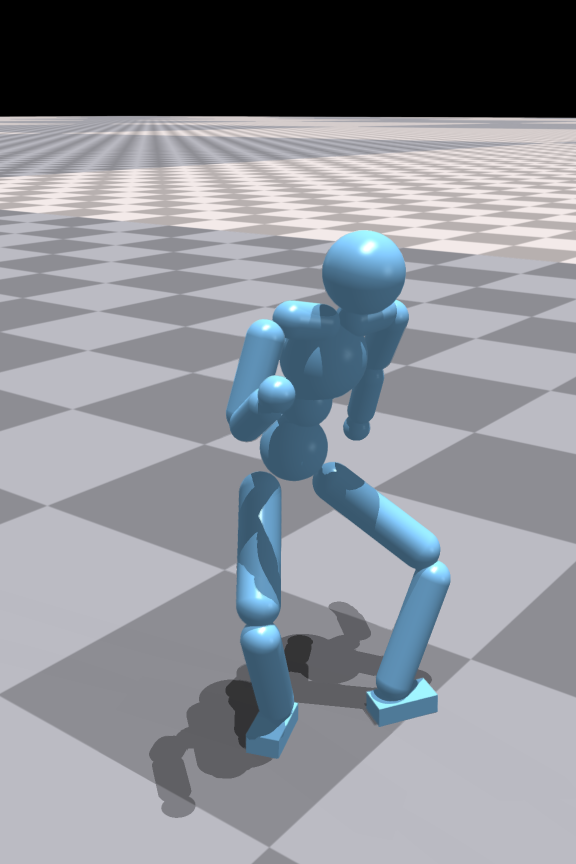} & 
\includegraphics[width=0.07\textwidth]{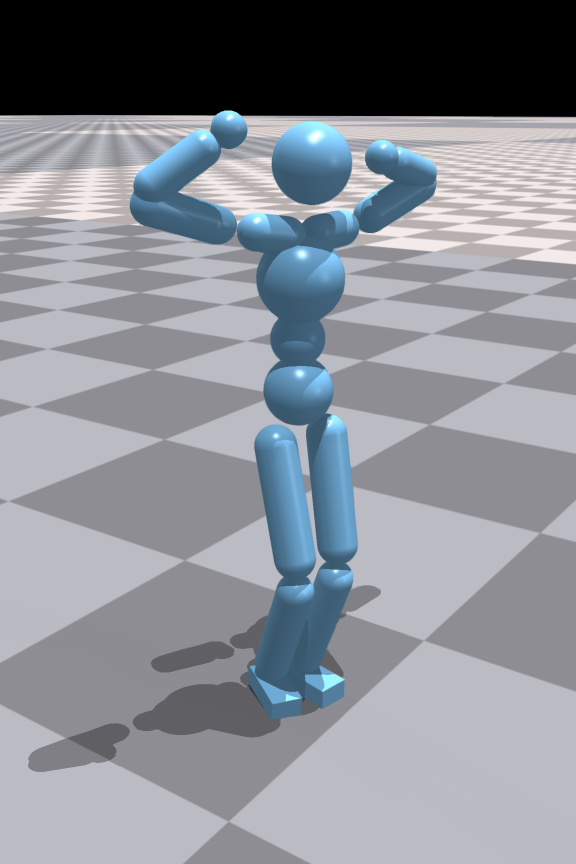} \\ 
\raisebox{-0.01\height}{\includegraphics[width=0.171\textwidth]{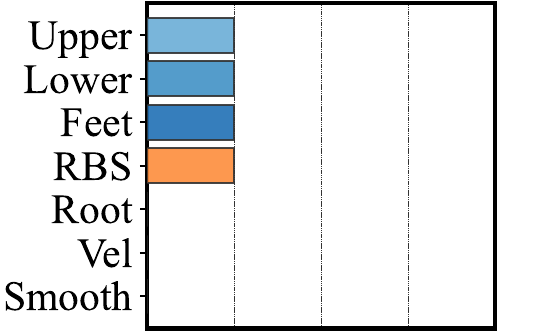}} & 
\includegraphics[width=0.07\textwidth]{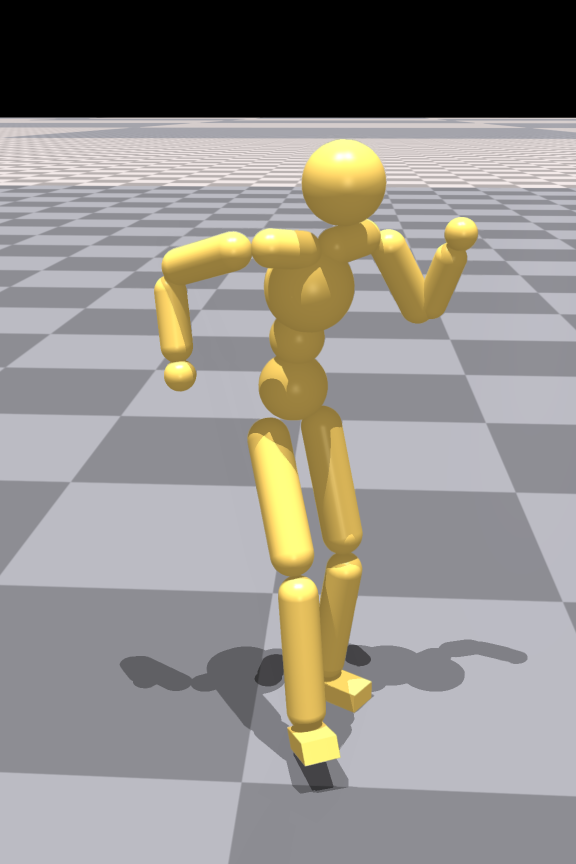} & 
\includegraphics[width=0.07\textwidth]{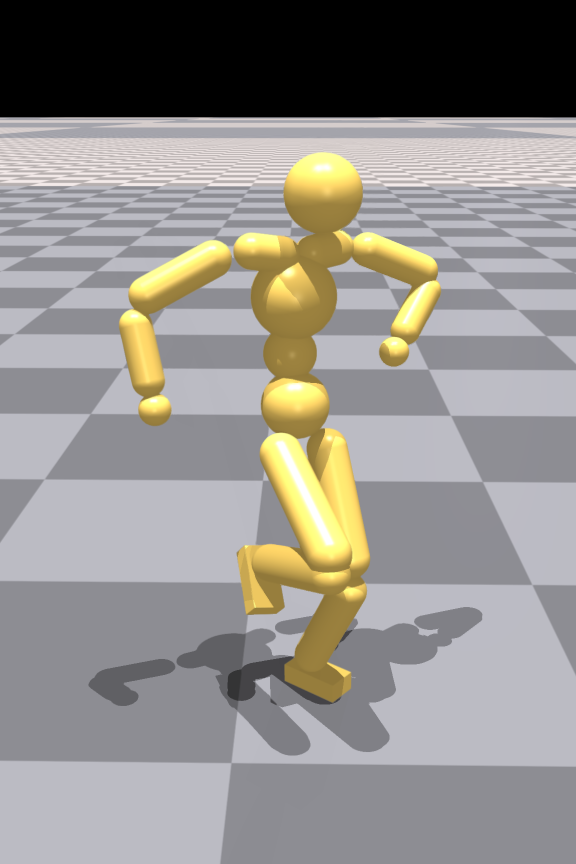} & 
\includegraphics[width=0.07\textwidth]{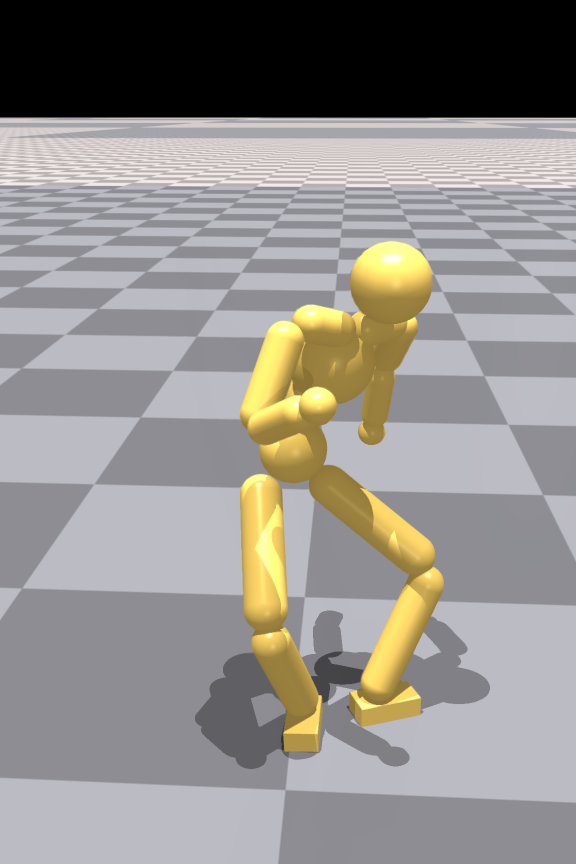} & 
\includegraphics[width=0.07\textwidth]{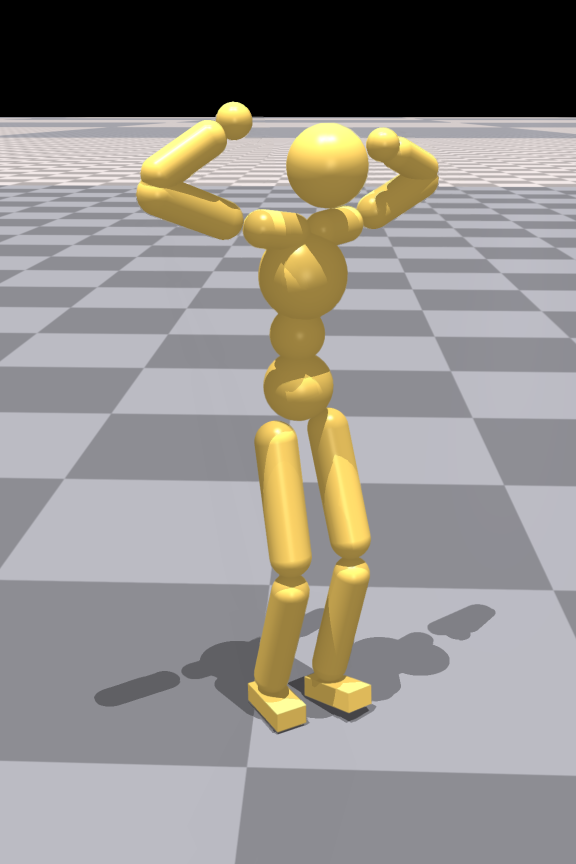} \\ 
\raisebox{-0.15\height}{\includegraphics[width=0.171\textwidth]{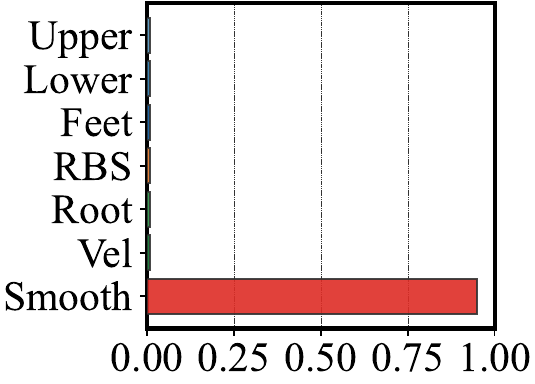}} &
\includegraphics[width=0.07\textwidth]{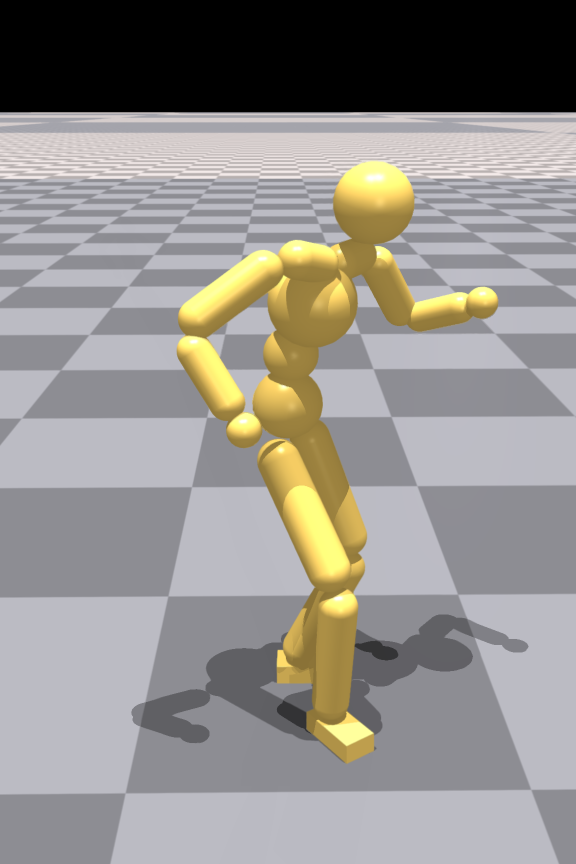} & 
\includegraphics[width=0.07\textwidth]{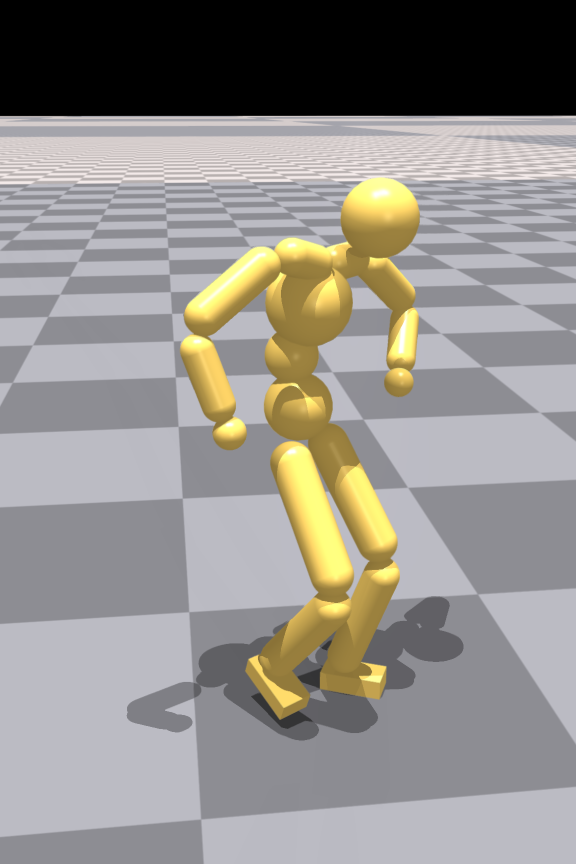} & 
\includegraphics[width=0.07\textwidth]{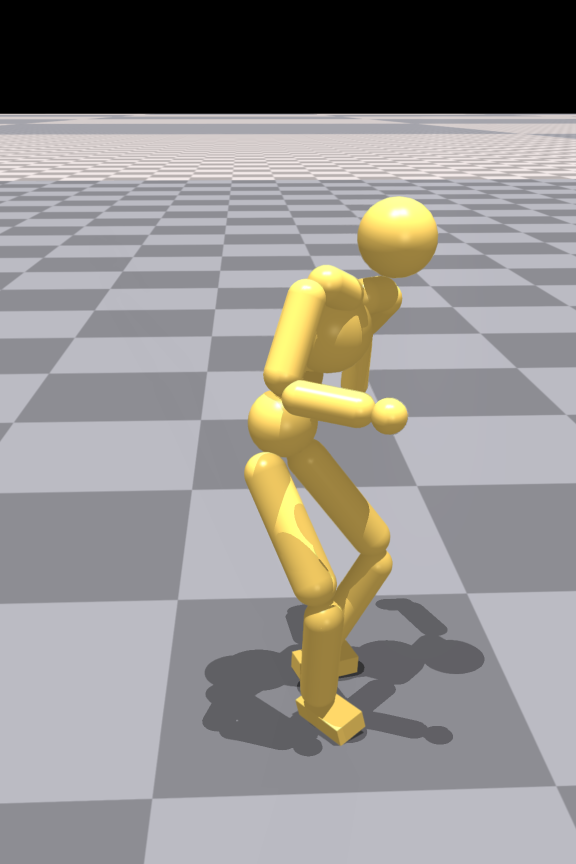} & 
\includegraphics[width=0.07\textwidth]{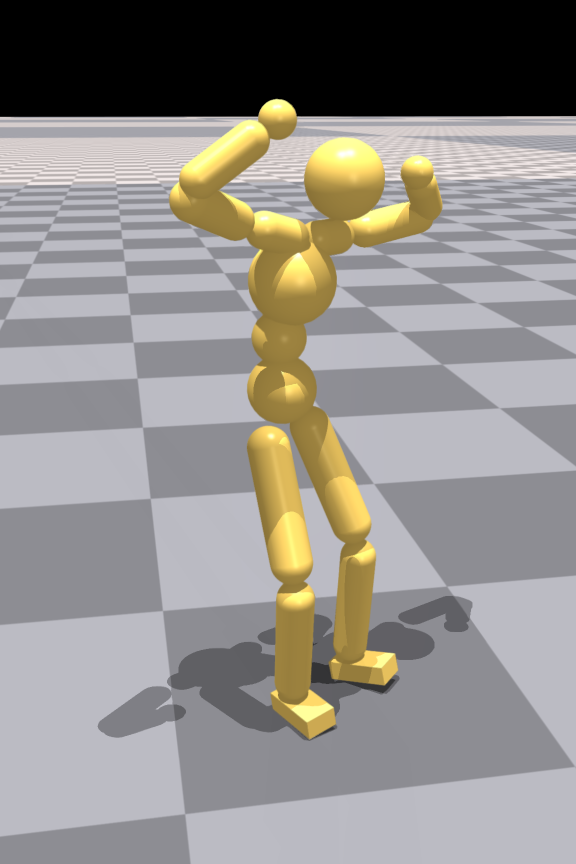} \\ 

\end{tabular}
}
\caption{\textbf{Weight Influence.} (Top) Kinematic reference. (Middle) Visual performance when prioritizing tracking reward weights. (Bottom) Visual performance when prioritizing smoothness reward terms.}
\label{fig:humanoid_dancing_through_weights}
\Description{Humanoid dancing through weights}
\end{figure}

\paragraph{Robot}

We use AMOR on a 20-DoF bipedal robot (Fig.~\ref{fig:teaser}), and compare a uniform weight distribution against manually-tuned weights on two challenging examples. 

We first consider a stationary dancing motion, and compare the simulated robot to the physical one. As seen in Fig.~\ref{fig:robot_weight_tuning_joints}, the real robot exhibits significantly higher jitter in the joint positions and velocities. By increasing the smoothness objective, we can reduce this jitter, and thereby also reduce the sim-to-real gap as shown in the figure. This experiment is highlighted in the supplementary video and best appreciated by focusing on the actuator sound. 
 
 In the second example, a double pirouette motion, shown in Fig.~\ref{fig:teaser} and the accompanying video, we demonstrate that tuning the reward weights on the real system enables the robot to perform physically demanding motions beyond the capabilities of the state-of-the-art fixed-weight VMP controller. To achieve this motion, we experimentally determined the impact of each reward term through on-the-fly weight adjustments directly on the robot. 
 
 We found that time-varying weights were required, to address particular challenges of different parts of the motion. For example, a high velocity weight was required during the initial part of the motion, to build up enough rotational speed for a stable pirouette. Conversely, a higher smoothing weight was required towards the end of the sequence, for a smoother transition out of the pirouette.

Tuning the weights for this dynamic motion took approximately 1 day of experimentation. Compared to the 5 days it takes to train the RL policy, it is clear that tuning with retraining would be infeasible.

\begin{figure*}
    \definecolor{color1}{HTML}{DC143C}
    \definecolor{color2}{HTML}{FFD700}
    \definecolor{color3}{HTML}{228B22} 
    \definecolor{color4}{HTML}{000080}
    \centering
    \setlength{\fboxsep}{0pt}  
    \setlength{\fboxrule}{1.75pt} 
    \renewcommand{\arraystretch}{0} 
    \setlength{\tabcolsep}{0pt}

    \begin{tabular}{@{}c@{}c@{}c@{}c@{}}

        \fcolorbox{color1!40}{white}{%
            \begin{tabular}{@{}c@{\hspace{1pt}}c@{\hspace{1pt}}c@{\hspace{1pt}}c@{}}
                \includegraphics[width=0.058\textwidth]{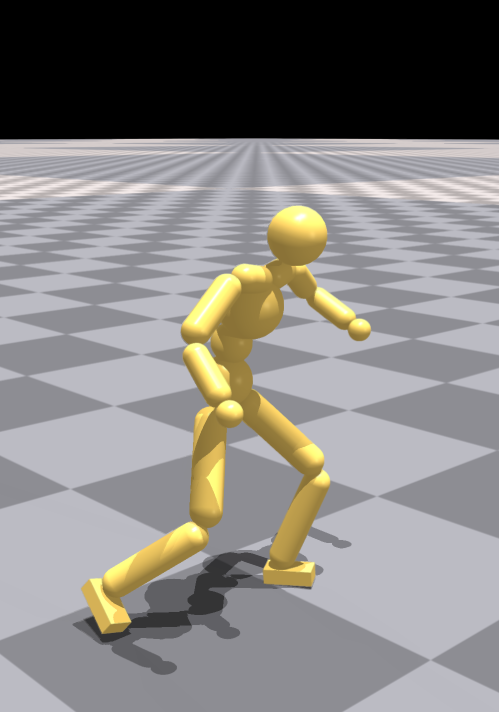} &
                \includegraphics[width=0.058\textwidth]{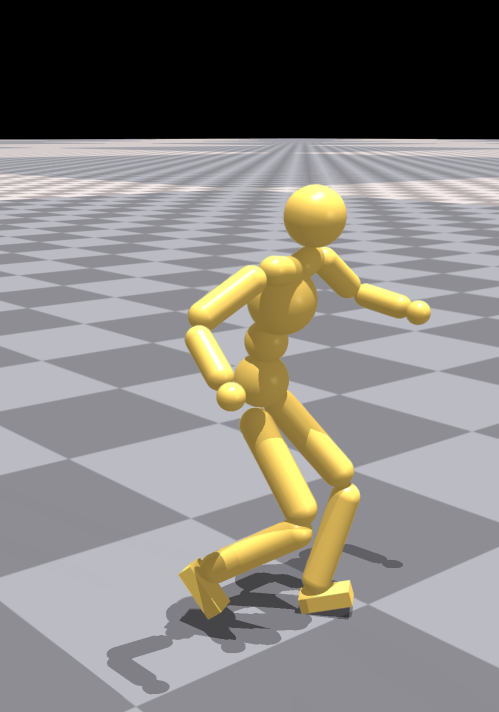} &
                \includegraphics[width=0.058\textwidth]{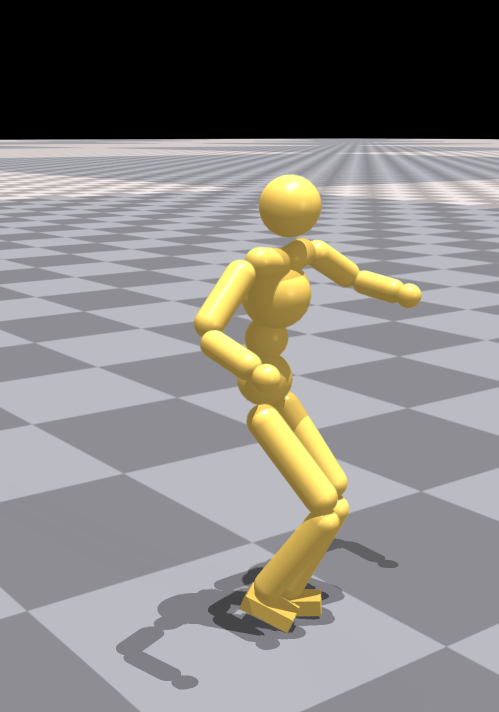} &
                \includegraphics[width=0.058\textwidth]{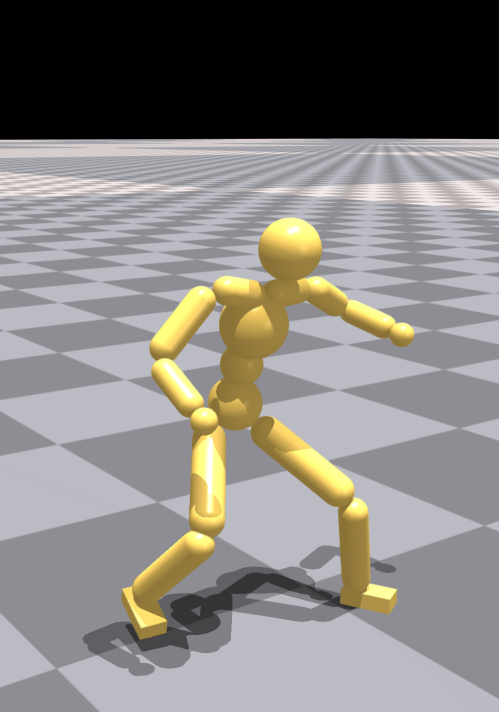}
            \end{tabular}%
        } & \hspace{0.25em}

        \fcolorbox{color2!40}{white}{%
            \begin{tabular}{@{}c@{\hspace{1pt}}c@{\hspace{1pt}}c@{\hspace{1pt}}c@{}}
                \includegraphics[width=0.058\textwidth]{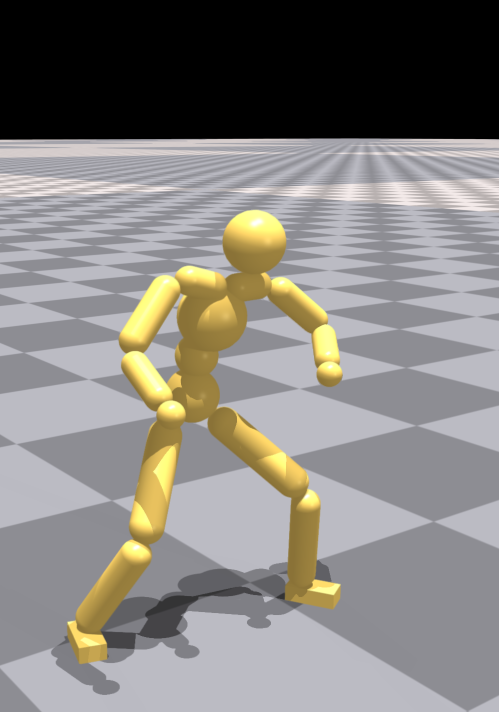} &
                \includegraphics[width=0.058\textwidth]{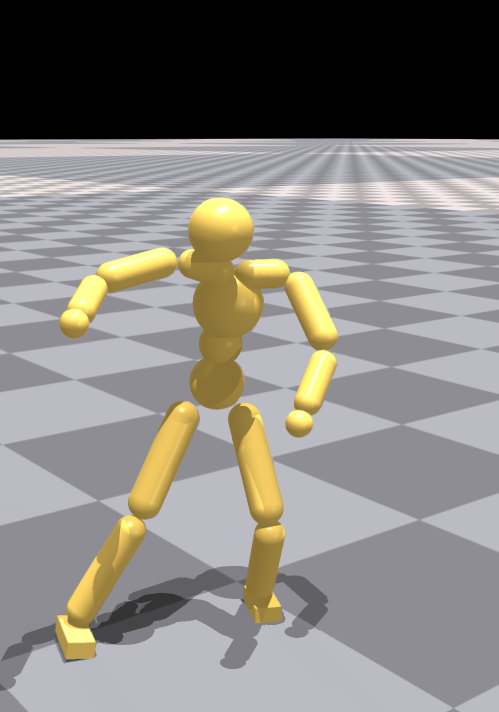} &
                \includegraphics[width=0.058\textwidth]{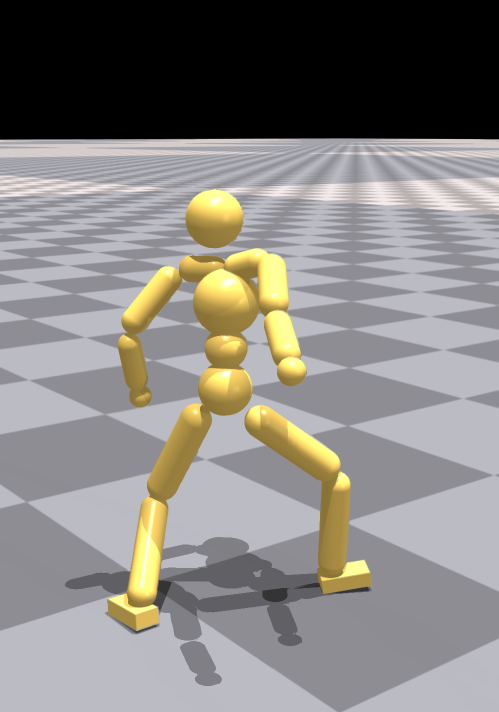} &
                \includegraphics[width=0.058\textwidth]{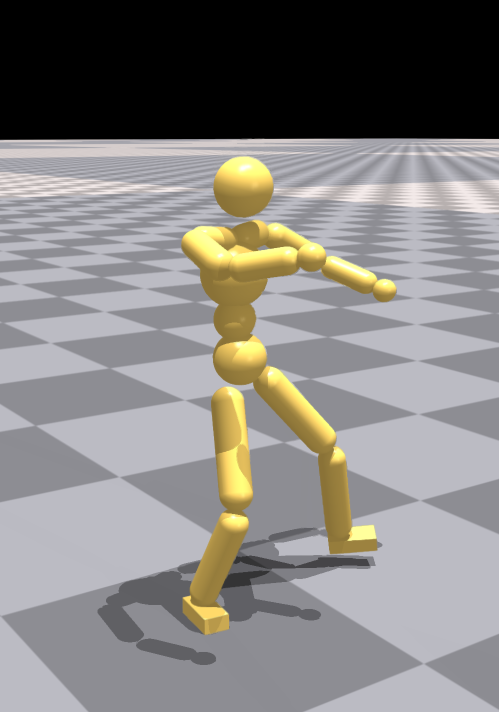}
            \end{tabular}%
        } & \hspace{0.25em}

        \fcolorbox{color3!40}{white}{%
            \begin{tabular}{@{}c@{\hspace{1pt}}c@{\hspace{1pt}}c@{\hspace{1pt}}c@{}}
                \includegraphics[width=0.058\textwidth]{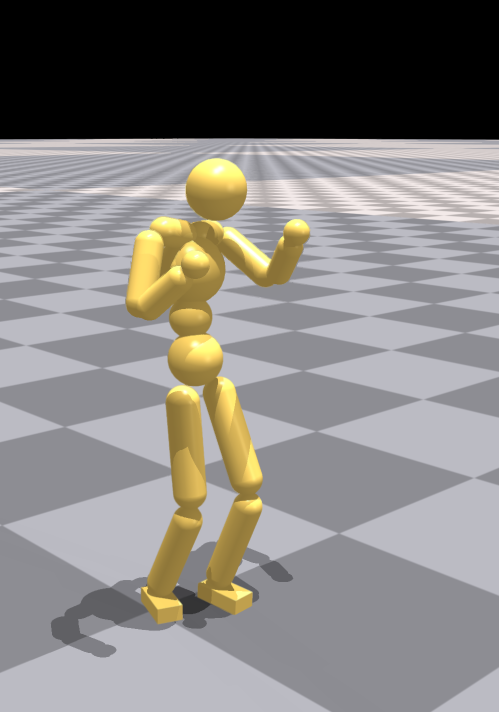} &
                \includegraphics[width=0.058\textwidth]{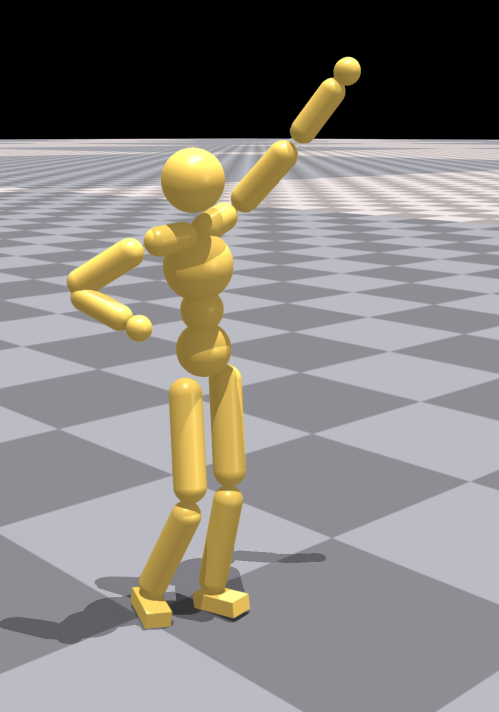} &
                \includegraphics[width=0.058\textwidth]{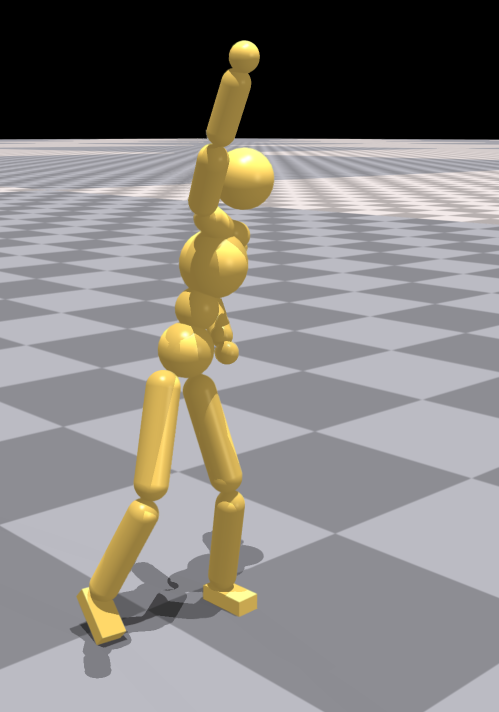} &
                \includegraphics[width=0.058\textwidth]{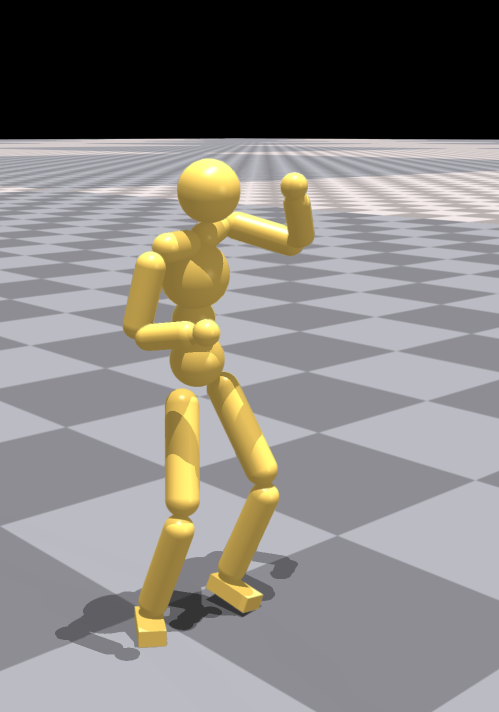}
            \end{tabular}%
        } & \hspace{0.25em}

        \fcolorbox{color4!40}{white}{%
            \begin{tabular}{@{}c@{\hspace{1pt}}c@{\hspace{1pt}}c@{\hspace{1pt}}c@{}}
                \includegraphics[width=0.058\textwidth]{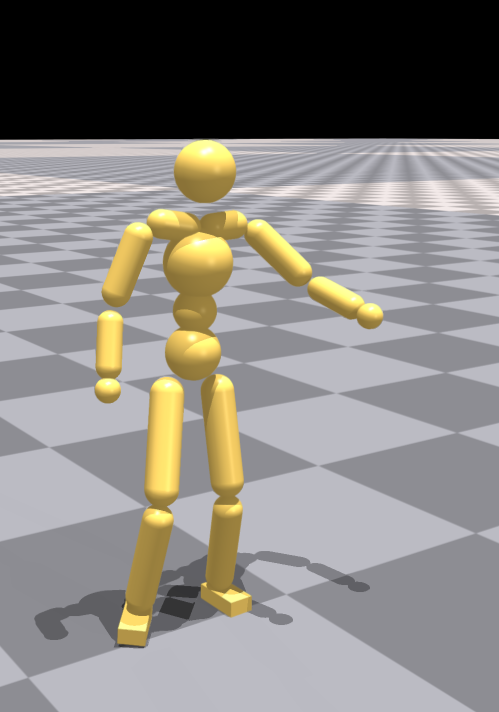} &
                \includegraphics[width=0.058\textwidth]{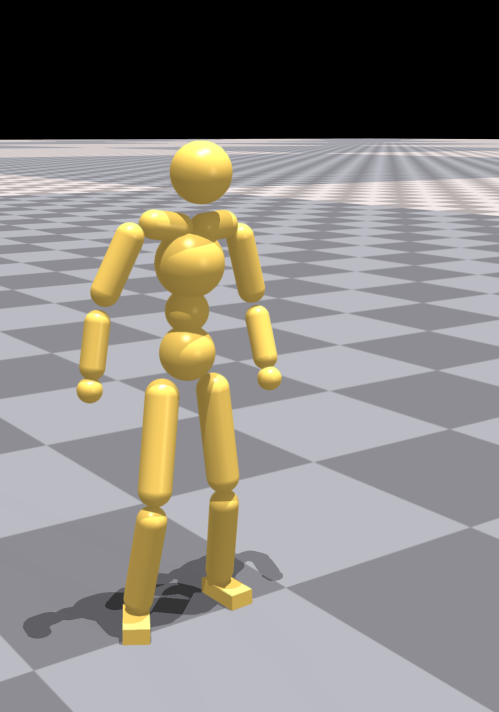} &
                \includegraphics[width=0.058\textwidth]{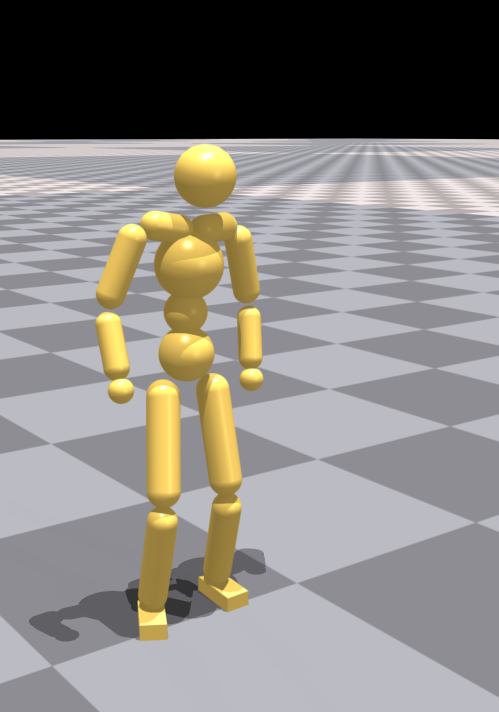} &
                \includegraphics[width=0.058\textwidth]{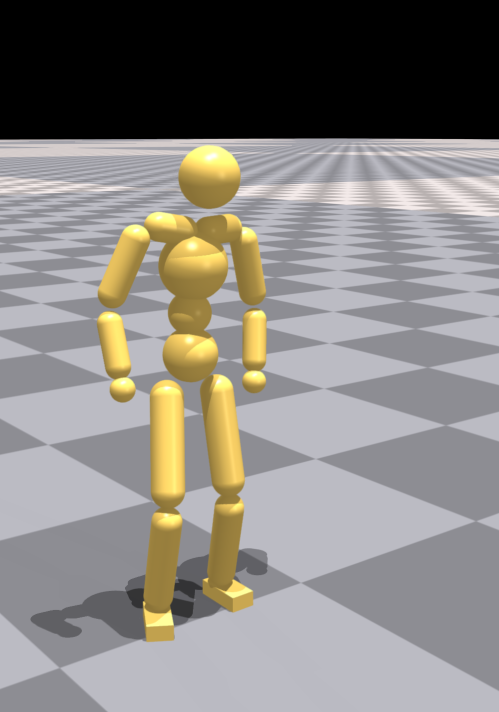}
            \end{tabular}%
        }

    \end{tabular}
    
    \vspace{0.25em}
    \includegraphics[width=0.99\linewidth]{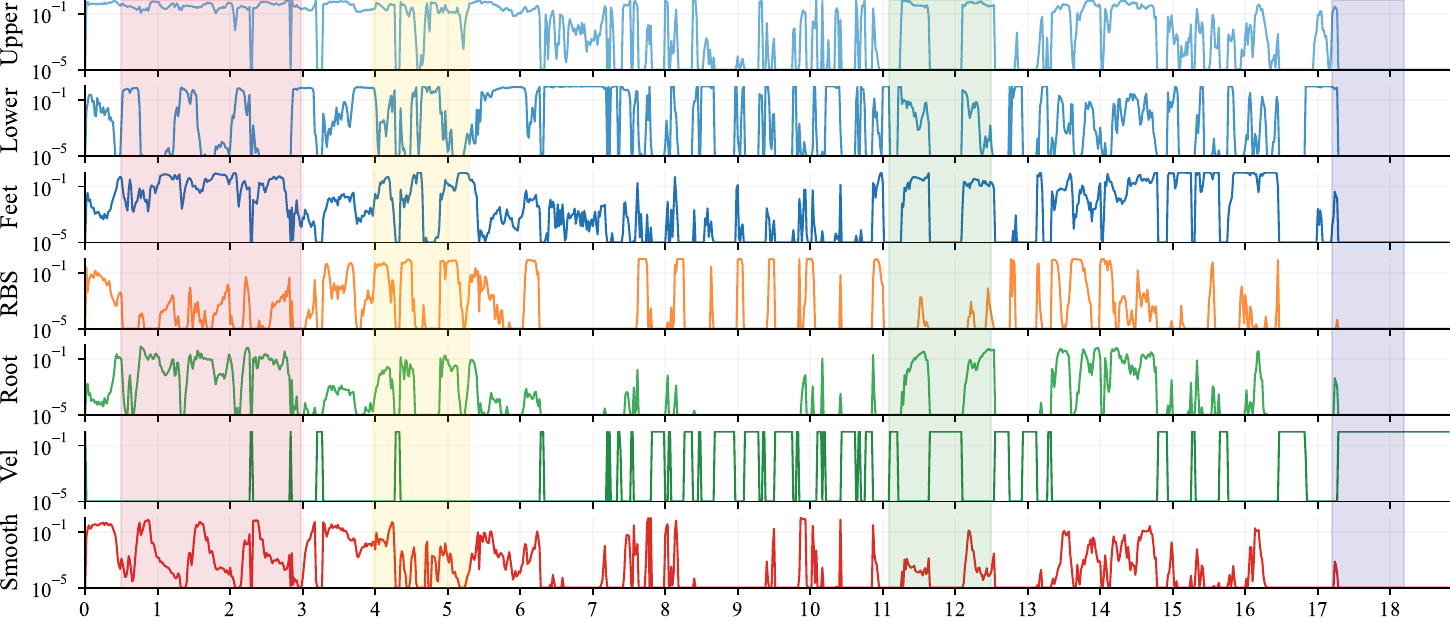}
    \caption{\textbf{Adaptive Weights.} Visualization of the behavior of AMOR's high-level policy $\pih(\w_t|\mathbf{s}_t,\mathbf{c_t})$ through different types of motion. The resulting reward weights $\w_t \in \Delta^m$ are plotted on a logarithmic scale; this better captures their multiplicative effect on the reward.}
    \label{fig:lima_meta_30}
    \Description{Lima high-level policy.}
\end{figure*}

\begin{figure*}
    \centering
    \includegraphics[width=0.99\linewidth]{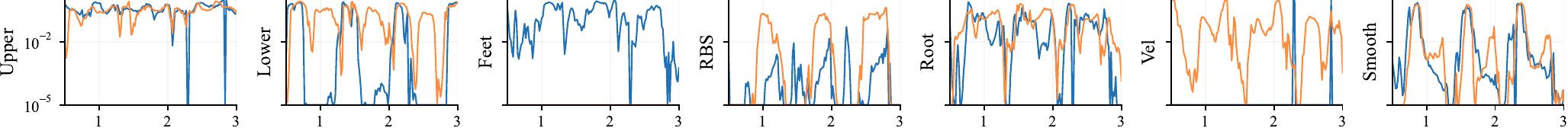}
    \caption{\textbf{Discriminator Window Size} Comparison between \textcolor{feet}{30-frame} and \textcolor{rbs}{2-frame} discriminator rewards on the HLP for the first interval of Fig.~\ref{fig:lima_meta_30}. The different discriminator window lengths lead to different reward weights.}
    \label{fig:meta_comparison_2_30_windows}
\end{figure*}

\subsection{High-Level Policy}

\paragraph{Hierarchical Weight Adjustment}

We now evaluate the behavior of the proposed HLP, in particular studying how the weights returned by the HLP vary over time and with the motion being performed. 
In Fig.~\ref{fig:lima_meta_30}, we show how the reward weights produced by $\pih(\w_t|\mathbf{s}_t,\mathbf{c_t})$ evolve as the humanoid transitions through walking, spinning, performing an upper-body motion, and standing still. 

During walking (red background), the HLP balances the feet and lower body coordination combined with smoothing to feather the steps of the character. The activations of the velocity term are generally short and intermittent; and can for example be seen to align with the start of a spinning motion (yellow window). The motion in the green window features two high punches in the air; this can be seen to correlate with two spikes in the tracking weights (blue curves). During the final part of the motion (purple background) the reference motion remains at rest; this can be seen to activate the velocity weight.

Interestingly, we observe that changing the window size $V$ of the observations for the discrimination produces different behaviors. Fig.~\ref{fig:meta_comparison_2_30_windows} shows a comparison between the 30-frame discriminator and a 2-frame discriminator. While it is difficult to draw conclusions, we can observe that the 2-frame version weighs the velocity term more heavily, while the 30-frame version instead places higher weights on tracking the lower body and feet.

\paragraph{Motion Tracking Performance}

We evaluate the HLP by comparing humanoid performance in simulation using HLP's weight predictions against equally-balanced fixed weights. First, we measure the mean absolute error (MAE) of joint positions relative to a kinematic reference, using 80'000 randomly-sampled 30-second motion episodes. For shorter motions, additional segments are appended to complete the episode. Second, we assess the logits of a discriminator $D(\mathbf{O}_t|\mathbf{z}_t)$ trained on 32M 30-frame windows $\mathbf{O}_t$ collected from simulations with HLP, fixed weights, and dataset transitions. The discriminator, trained to distinguish between simulated and dataset transitions, is evaluated on 8M windows. Note that this discriminator is trained separately for this evaluation, and is not the discriminator used to train the HLP. Results in Tab.~\ref{tab:comparision_hlp_amor_fixed} show significant improvements in both explicit MAE and implicit assessment through logits when using HLP over fixed weights. This highlights that the hierarchical approach is indeed able to leverage the adaptability of the AMOR policy.

\begin{table}
\caption{\textbf{Motion Tracking Performance.} Comparison of AMOR's performance on the simulated humanoid using HLP against equally balanced fixed weights is measured both explicitly, using the mean absolute error (MAE) of the character's joint pose in degrees, and implicitly, through the logits of a discriminator.}
\label{tab:comparision_hlp_amor_fixed}
\centering
\begin{footnotesize}
\begin{tabular}{l rr}
\toprule
Weights \hspace{1.5cm} & MAE $\downarrow$ & Logits $\uparrow$\\
\midrule
Uniform Weights           & $10.02$ & $-18.02$ \\
HLP Prediction         & $\mathbf{9.55}$ & $\mathbf{-15.48}$ \\
\bottomrule
\end{tabular}
\end{footnotesize}
\end{table}

\section{Conclusions}

We have introduced AMOR, a policy conditioned on context and a linear combination of reward weights trained with MORL. We have shown that a multi-objective approach scales to the problem complexity found in motion tracking for physics-based characters and robots. By leveraging context conditioning, AMOR can successfully track a wide variety of motions under different reward weights.
AMOR allows for on-the-fly adjustment of reward weights after training, unlocking new possibilities, some of which we have explored in this work. While our work targets character control, we expect the adaptive weight tuning enabled by AMOR to be applicable in other domains as well.

We demonstrated the use of AMOR for sim-to-real transfer by tuning rewards to reduce the sim-to-real gap, leveraging on-the-fly reward adjustments and time-varying rewards to push the boundaries of dynamic robot motions.

Moreover, we explored a hierarchical extension to AMOR, by training a HLP which dynamically adjusts reward weights over time to closely match a given motion reference, using a discriminator reward. An interesting direction for future work is to explore alternative higher-level task-based rewards for the HLP.

Another potential application is automated fine-tuning using robot-in-the-loop online reinforcement learning (ORL)~\cite{wu2022daydreamerworldmodelsphysical}. Using AMOR as a low-level policy could reduce training time, and mitigate the high failure rates observed early in training.

Interestingly, some performance gap is apparent between regular PPO and MOPPO due to the increased problem complexity. In future work, this could potentially be reduced through large-scale training until convergence, or through alternative MORL techniques, e.g., by employing different strategies for sampling weight vectors during training~\cite{alegre2023sample}.

\bibliographystyle{ACM-Reference-Format}
\bibliography{references}


\begin{thebibliography}{83}


\ifx \showCODEN    \undefined \def \showCODEN     #1{\unskip}     \fi
\ifx \showISBNx    \undefined \def \showISBNx     #1{\unskip}     \fi
\ifx \showISBNxiii \undefined \def \showISBNxiii  #1{\unskip}     \fi
\ifx \showISSN     \undefined \def \showISSN      #1{\unskip}     \fi
\ifx \showLCCN     \undefined \def \showLCCN      #1{\unskip}     \fi
\ifx \shownote     \undefined \def \shownote      #1{#1}          \fi
\ifx \showarticletitle \undefined \def \showarticletitle #1{#1}   \fi
\ifx \showURL      \undefined \def \showURL       {\relax}        \fi
\providecommand\bibfield[2]{#2}
\providecommand\bibinfo[2]{#2}
\providecommand\natexlab[1]{#1}
\providecommand\showeprint[2][]{arXiv:#2}

\bibitem[Al~Borno et~al\mbox{.}(2013)]%
        {borno2013trajectory}
\bibfield{author}{\bibinfo{person}{Mazen Al~Borno}, \bibinfo{person}{Martin de Lasa}, {and} \bibinfo{person}{Aaron Hertzmann}.} \bibinfo{year}{2013}\natexlab{}.
\newblock \showarticletitle{Trajectory Optimization for Full-Body Movements with Complex Contacts}.
\newblock \bibinfo{journal}{\emph{IEEE Transactions on Visualization and Computer Graphics}} \bibinfo{volume}{19}, \bibinfo{number}{8} (\bibinfo{year}{2013}), \bibinfo{pages}{1405--1414}.
\newblock
\href{https://doi.org/10.1109/TVCG.2012.325}{doi:\nolinkurl{10.1109/TVCG.2012.325}}


\bibitem[Alegre et~al\mbox{.}(2022)]%
        {alegre2022}
\bibfield{author}{\bibinfo{person}{Lucas~N. Alegre}, \bibinfo{person}{Ana L.~C. Bazzan}, {and} \bibinfo{person}{Bruno~C. {da Silva}}.} \bibinfo{year}{2022}\natexlab{}.
\newblock \showarticletitle{Optimistic Linear Support and Successor Features as a Basis for Optimal Policy Transfer}. In \bibinfo{booktitle}{\emph{Proceedings of the 39th International Conference on Machine Learning}} \emph{(\bibinfo{series}{Proceedings of Machine Learning Research}, Vol.~\bibinfo{volume}{162})}, \bibfield{editor}{\bibinfo{person}{Kamalika Chaudhuri}, \bibinfo{person}{Stefanie Jegelka}, \bibinfo{person}{Le~Song}, \bibinfo{person}{Csaba Szepesvari}, \bibinfo{person}{Gang Niu}, {and} \bibinfo{person}{Sivan Sabato}} (Eds.). \bibinfo{publisher}{PMLR}, \bibinfo{pages}{394--413}.
\newblock


\bibitem[Alegre et~al\mbox{.}(2023)]%
        {alegre2023sample}
\bibfield{author}{\bibinfo{person}{Lucas~N. Alegre}, \bibinfo{person}{Ana L.~C. Bazzan}, \bibinfo{person}{Diederik~M. Roijers}, \bibinfo{person}{Ann Now{\'e}}, {and} \bibinfo{person}{Bruno~C. da Silva}.} \bibinfo{year}{2023}\natexlab{}.
\newblock \showarticletitle{Sample-Efficient Multi-Objective Learning via Generalized Policy Improvement Prioritization}. In \bibinfo{booktitle}{\emph{Proceedings of the 2023 International Conference on Autonomous Agents and Multiagent Systems}} (London, United Kingdom) \emph{(\bibinfo{series}{AAMAS '23})}. \bibinfo{publisher}{International Foundation for Autonomous Agents and Multiagent Systems}, \bibinfo{address}{Richland, SC}, \bibinfo{pages}{2003--2012}.
\newblock
\showISBNx{9781450394321}


\bibitem[Andrychowicz et~al\mbox{.}(2021)]%
        {andrychowicz2021what}
\bibfield{author}{\bibinfo{person}{Marcin Andrychowicz}, \bibinfo{person}{Anton Raichuk}, \bibinfo{person}{Piotr Sta{\'n}czyk}, \bibinfo{person}{Manu Orsini}, \bibinfo{person}{Sertan Girgin}, \bibinfo{person}{Rapha{\"e}l Marinier}, \bibinfo{person}{Leonard Hussenot}, \bibinfo{person}{Matthieu Geist}, \bibinfo{person}{Olivier Pietquin}, \bibinfo{person}{Marcin Michalski}, \bibinfo{person}{Sylvain Gelly}, {and} \bibinfo{person}{Olivier Bachem}.} \bibinfo{year}{2021}\natexlab{}.
\newblock \showarticletitle{What Matters for On-Policy Deep Actor-Critic Methods? A Large-Scale Study}. In \bibinfo{booktitle}{\emph{Proceedings of the Ninth International Conference on Learning Representations}}.
\newblock
\urldef\tempurl%
\url{https://openreview.net/forum?id=nIAxjsniDzg}
\showURL{%
\tempurl}


\bibitem[Barreto et~al\mbox{.}(2019)]%
        {barreto2019option}
\bibfield{author}{\bibinfo{person}{Andr\'{e} Barreto}, \bibinfo{person}{Diana Borsa}, \bibinfo{person}{Shaobo Hou}, \bibinfo{person}{Gheorghe Comanici}, \bibinfo{person}{Eser Ayg\"{u}n}, \bibinfo{person}{Philippe Hamel}, \bibinfo{person}{Daniel Toyama}, \bibinfo{person}{Jonathan Hunt}, \bibinfo{person}{Shibl Mourad}, \bibinfo{person}{David Silver}, {and} \bibinfo{person}{Doina Precup}.} \bibinfo{year}{2019}\natexlab{}.
\newblock \showarticletitle{The option keyboard combining skills in reinforcement learning}. In \bibinfo{booktitle}{\emph{Proceedings of the 33rd International Conference on Neural Information Processing Systems}}. \bibinfo{publisher}{Curran Associates Inc.}, \bibinfo{address}{Red Hook, NY, USA}, Article \bibinfo{articleno}{1169}, \bibinfo{numpages}{11}~pages.
\newblock


\bibitem[Barreto et~al\mbox{.}(2017)]%
        {barreto2017successor}
\bibfield{author}{\bibinfo{person}{Andr{\'e} Barreto}, \bibinfo{person}{Will Dabney}, \bibinfo{person}{R{\'e}mi Munos}, \bibinfo{person}{Jonathan~J Hunt}, \bibinfo{person}{Tom Schaul}, \bibinfo{person}{Hado~P van Hasselt}, {and} \bibinfo{person}{David Silver}.} \bibinfo{year}{2017}\natexlab{}.
\newblock \showarticletitle{Successor features for transfer in reinforcement learning}.
\newblock \bibinfo{journal}{\emph{Advances in neural information processing systems}}  \bibinfo{volume}{30} (\bibinfo{year}{2017}).
\newblock


\bibitem[Bergamin et~al\mbox{.}(2019)]%
        {bergamin2019drecon}
\bibfield{author}{\bibinfo{person}{Kevin Bergamin}, \bibinfo{person}{Simon Clavet}, \bibinfo{person}{Daniel Holden}, {and} \bibinfo{person}{James~Richard Forbes}.} \bibinfo{year}{2019}\natexlab{}.
\newblock \showarticletitle{DReCon: data-driven responsive control of physics-based characters}.
\newblock \bibinfo{journal}{\emph{ACM Trans. Graph.}} \bibinfo{volume}{38}, \bibinfo{number}{6}, Article \bibinfo{articleno}{206} (\bibinfo{date}{Nov.} \bibinfo{year}{2019}), \bibinfo{numpages}{11}~pages.
\newblock
\showISSN{0730-0301}
\href{https://doi.org/10.1145/3355089.3356536}{doi:\nolinkurl{10.1145/3355089.3356536}}


\bibitem[Cheng et~al\mbox{.}(2024)]%
        {cheng2024expressive}
\bibfield{author}{\bibinfo{person}{Xuxin Cheng}, \bibinfo{person}{Yandong Ji}, \bibinfo{person}{Junming Chen}, \bibinfo{person}{Ruihan Yang}, \bibinfo{person}{Ge Yang}, {and} \bibinfo{person}{Xiaolong Wang}.} \bibinfo{year}{2024}\natexlab{}.
\newblock \showarticletitle{Expressive whole-body control for humanoid robots}.
\newblock \bibinfo{journal}{\emph{arXiv preprint arXiv:2402.16796}} (\bibinfo{year}{2024}).
\newblock


\bibitem[Chentanez et~al\mbox{.}(2018)]%
        {chentanez2018physics}
\bibfield{author}{\bibinfo{person}{Nuttapong Chentanez}, \bibinfo{person}{Matthias M\"{u}ller}, \bibinfo{person}{Miles Macklin}, \bibinfo{person}{Viktor Makoviychuk}, {and} \bibinfo{person}{Stefan Jeschke}.} \bibinfo{year}{2018}\natexlab{}.
\newblock \showarticletitle{Physics-based motion capture imitation with deep reinforcement learning}. In \bibinfo{booktitle}{\emph{Proceedings of the 11th ACM SIGGRAPH Conference on Motion, Interaction and Games}} (Limassol, Cyprus) \emph{(\bibinfo{series}{MIG '18})}. \bibinfo{publisher}{Association for Computing Machinery}, \bibinfo{address}{New York, NY, USA}, Article \bibinfo{articleno}{1}, \bibinfo{numpages}{10}~pages.
\newblock
\showISBNx{9781450360159}
\href{https://doi.org/10.1145/3274247.3274506}{doi:\nolinkurl{10.1145/3274247.3274506}}


\bibitem[Clevert et~al\mbox{.}(2016)]%
        {clevert2016fast}
\bibfield{author}{\bibinfo{person}{Djork{-}Arn{\'{e}} Clevert}, \bibinfo{person}{Thomas Unterthiner}, {and} \bibinfo{person}{Sepp Hochreiter}.} \bibinfo{year}{2016}\natexlab{}.
\newblock \showarticletitle{Fast and Accurate Deep Network Learning by Exponential Linear Units (ELUs)}. In \bibinfo{booktitle}{\emph{4th International Conference on Learning Representations, {ICLR} 2016, San Juan, Puerto Rico, May 2-4, 2016, Conference Track Proceedings}}, \bibfield{editor}{\bibinfo{person}{Yoshua Bengio} {and} \bibinfo{person}{Yann LeCun}} (Eds.).
\newblock
\urldef\tempurl%
\url{http://arxiv.org/abs/1511.07289}
\showURL{%
\tempurl}


\bibitem[CMU(2001)]%
        {cmu}
\bibfield{author}{\bibinfo{person}{CMU}.} \bibinfo{year}{2001}\natexlab{}.
\newblock \bibinfo{booktitle}{\emph{CMU Graphics Lab Motion Capture Database}}.
\newblock
\urldef\tempurl%
\url{http://mocap.cs.cmu.edu/}
\showURL{%
\tempurl}


\bibitem[Coros et~al\mbox{.}(2010)]%
        {coros2010generalized}
\bibfield{author}{\bibinfo{person}{Stelian Coros}, \bibinfo{person}{Philippe Beaudoin}, {and} \bibinfo{person}{Michiel van~de Panne}.} \bibinfo{year}{2010}\natexlab{}.
\newblock \showarticletitle{Generalized biped walking control}.
\newblock \bibinfo{journal}{\emph{ACM Trans. Graph.}} \bibinfo{volume}{29}, \bibinfo{number}{4}, Article \bibinfo{articleno}{130} (\bibinfo{date}{July} \bibinfo{year}{2010}), \bibinfo{numpages}{9}~pages.
\newblock
\showISSN{0730-0301}
\href{https://doi.org/10.1145/1778765.1781156}{doi:\nolinkurl{10.1145/1778765.1781156}}


\bibitem[Dayan(1993)]%
        {dayan1993successor}
\bibfield{author}{\bibinfo{person}{Peter Dayan}.} \bibinfo{year}{1993}\natexlab{}.
\newblock \showarticletitle{Improving Generalization for Temporal Difference Learning: The Successor Representation}.
\newblock \bibinfo{journal}{\emph{Neural Computation}} \bibinfo{volume}{5}, \bibinfo{number}{4} (\bibinfo{year}{1993}), \bibinfo{pages}{613--624}.
\newblock
\href{https://doi.org/10.1162/neco.1993.5.4.613}{doi:\nolinkurl{10.1162/neco.1993.5.4.613}}


\bibitem[Dou et~al\mbox{.}(2023)]%
        {dou2023case}
\bibfield{author}{\bibinfo{person}{Zhiyang Dou}, \bibinfo{person}{Xuelin Chen}, \bibinfo{person}{Qingnan Fan}, \bibinfo{person}{Taku Komura}, {and} \bibinfo{person}{Wenping Wang}.} \bibinfo{year}{2023}\natexlab{}.
\newblock \showarticletitle{C·ASE: Learning Conditional Adversarial Skill Embeddings for Physics-based Characters}. In \bibinfo{booktitle}{\emph{SIGGRAPH Asia 2023 Conference Papers}} (Sydney, NSW, Australia) \emph{(\bibinfo{series}{SA '23})}. \bibinfo{publisher}{Association for Computing Machinery}, \bibinfo{address}{New York, NY, USA}, Article \bibinfo{articleno}{2}, \bibinfo{numpages}{11}~pages.
\newblock
\showISBNx{9798400703157}
\href{https://doi.org/10.1145/3610548.3618205}{doi:\nolinkurl{10.1145/3610548.3618205}}


\bibitem[Felten et~al\mbox{.}(2023)]%
        {felten2023toolkit}
\bibfield{author}{\bibinfo{person}{Florian Felten}, \bibinfo{person}{Lucas~N. Alegre}, \bibinfo{person}{Ann Now{\'e}}, \bibinfo{person}{Ana L.~C. Bazzan}, \bibinfo{person}{El-Ghazali Talbi}, \bibinfo{person}{Gr{\'e}goire Danoy}, {and} \bibinfo{person}{Bruno~C. da Silva}.} \bibinfo{year}{2023}\natexlab{}.
\newblock \showarticletitle{A Toolkit for Reliable Benchmarking and Research in Multi-Objective Reinforcement Learning}. In \bibinfo{booktitle}{\emph{Advances in Neural Information Processing Systems}} (New Orleans, USA), Vol.~\bibinfo{volume}{36}.
\newblock


\bibitem[Felten et~al\mbox{.}(2024)]%
        {felten+2024}
\bibfield{author}{\bibinfo{person}{Florian Felten}, \bibinfo{person}{El-Ghazali Talbi}, {and} \bibinfo{person}{Gr{\'e}goire Danoy}.} \bibinfo{year}{2024}\natexlab{}.
\newblock \showarticletitle{Multi-Objective Reinforcement Learning based on Decomposition: A taxonomy and framework}.
\newblock \bibinfo{journal}{\emph{Journal of Artificial Intelligence Research}}  \bibinfo{volume}{79} (\bibinfo{year}{2024}), \bibinfo{pages}{679--723}.
\newblock


\bibitem[Fussell et~al\mbox{.}(2021)]%
        {fussell2021supertrack}
\bibfield{author}{\bibinfo{person}{Levi Fussell}, \bibinfo{person}{Kevin Bergamin}, {and} \bibinfo{person}{Daniel Holden}.} \bibinfo{year}{2021}\natexlab{}.
\newblock \showarticletitle{SuperTrack: motion tracking for physically simulated characters using supervised learning}.
\newblock \bibinfo{journal}{\emph{ACM Trans. Graph.}} \bibinfo{volume}{40}, \bibinfo{number}{6}, Article \bibinfo{articleno}{197} (\bibinfo{date}{Dec.} \bibinfo{year}{2021}), \bibinfo{numpages}{13}~pages.
\newblock
\showISSN{0730-0301}
\href{https://doi.org/10.1145/3478513.3480527}{doi:\nolinkurl{10.1145/3478513.3480527}}


\bibitem[Gehring et~al\mbox{.}(2023)]%
        {gehring2023leveraging}
\bibfield{author}{\bibinfo{person}{Jonas Gehring}, \bibinfo{person}{Deepak Gopinath}, \bibinfo{person}{Jungdam Won}, \bibinfo{person}{Andreas Krause}, \bibinfo{person}{Gabriel Synnaeve}, {and} \bibinfo{person}{Nicolas Usunier}.} \bibinfo{year}{2023}\natexlab{}.
\newblock \showarticletitle{Leveraging Demonstrations with Latent Space Priors}.
\newblock \bibinfo{journal}{\emph{Transactions on Machine Learning Research}} (\bibinfo{year}{2023}).
\newblock
\showISSN{2835-8856}
\urldef\tempurl%
\url{https://openreview.net/forum?id=OzGIu4T4Cz}
\showURL{%
\tempurl}


\bibitem[Goodfellow et~al\mbox{.}(2014)]%
        {goodfellow2014generative}
\bibfield{author}{\bibinfo{person}{Ian Goodfellow}, \bibinfo{person}{Jean Pouget-Abadie}, \bibinfo{person}{Mehdi Mirza}, \bibinfo{person}{Bing Xu}, \bibinfo{person}{David Warde-Farley}, \bibinfo{person}{Sherjil Ozair}, \bibinfo{person}{Aaron Courville}, {and} \bibinfo{person}{Yoshua Bengio}.} \bibinfo{year}{2014}\natexlab{}.
\newblock \showarticletitle{Generative adversarial nets}.
\newblock \bibinfo{journal}{\emph{Advances in neural information processing systems}}  \bibinfo{volume}{27} (\bibinfo{year}{2014}).
\newblock


\bibitem[Grandia et~al\mbox{.}(2024)]%
        {grandia2024design}
\bibfield{author}{\bibinfo{person}{Ruben Grandia}, \bibinfo{person}{Espen Knoop}, \bibinfo{person}{Michael~A Hopkins}, \bibinfo{person}{Georg Wiedebach}, \bibinfo{person}{Jared Bishop}, \bibinfo{person}{Steven Pickles}, \bibinfo{person}{David M{\"u}ller}, {and} \bibinfo{person}{Moritz B{\"a}cher}.} \bibinfo{year}{2024}\natexlab{}.
\newblock \showarticletitle{Design and control of a bipedal robotic character}. In \bibinfo{booktitle}{\emph{Proceedings of Robotics: Science and Systems}}.
\newblock


\bibitem[Gu et~al\mbox{.}(2025)]%
        {gu2025humanoid}
\bibfield{author}{\bibinfo{person}{Zhaoyuan Gu}, \bibinfo{person}{Junheng Li}, \bibinfo{person}{Wenlan Shen}, \bibinfo{person}{Wenhao Yu}, \bibinfo{person}{Zhaoming Xie}, \bibinfo{person}{Stephen McCrory}, \bibinfo{person}{Xianyi Cheng}, \bibinfo{person}{Abdulaziz Shamsah}, \bibinfo{person}{Robert Griffin}, \bibinfo{person}{C~Karen Liu}, {et~al\mbox{.}}} \bibinfo{year}{2025}\natexlab{}.
\newblock \showarticletitle{Humanoid Locomotion and Manipulation: Current Progress and Challenges in Control, Planning, and Learning}.
\newblock \bibinfo{journal}{\emph{arXiv preprint arXiv:2501.02116}} (\bibinfo{year}{2025}).
\newblock


\bibitem[Ha et~al\mbox{.}(2024)]%
        {ha2024learning}
\bibfield{author}{\bibinfo{person}{Sehoon Ha}, \bibinfo{person}{Joonho Lee}, \bibinfo{person}{Michiel van~de Panne}, \bibinfo{person}{Zhaoming Xie}, \bibinfo{person}{Wenhao Yu}, {and} \bibinfo{person}{Majid Khadiv}.} \bibinfo{year}{2024}\natexlab{}.
\newblock \showarticletitle{Learning-based legged locomotion; state of the art and future perspectives}.
\newblock \bibinfo{journal}{\emph{arXiv preprint arXiv:2406.01152}} (\bibinfo{year}{2024}).
\newblock


\bibitem[H\"{a}m\"{a}l\"{a}inen et~al\mbox{.}(2015)]%
        {hamalainen2015mpc}
\bibfield{author}{\bibinfo{person}{Perttu H\"{a}m\"{a}l\"{a}inen}, \bibinfo{person}{Joose Rajam\"{a}ki}, {and} \bibinfo{person}{C.~Karen Liu}.} \bibinfo{year}{2015}\natexlab{}.
\newblock \showarticletitle{Online control of simulated humanoids using particle belief propagation}.
\newblock \bibinfo{journal}{\emph{ACM Trans. Graph.}} \bibinfo{volume}{34}, \bibinfo{number}{4}, Article \bibinfo{articleno}{81} (\bibinfo{date}{July} \bibinfo{year}{2015}), \bibinfo{numpages}{13}~pages.
\newblock
\showISSN{0730-0301}
\href{https://doi.org/10.1145/2767002}{doi:\nolinkurl{10.1145/2767002}}


\bibitem[Hasenclever et~al\mbox{.}(2020)]%
        {hasenclever2020comic}
\bibfield{author}{\bibinfo{person}{Leonard Hasenclever}, \bibinfo{person}{Fabio Pardo}, \bibinfo{person}{Raia Hadsell}, \bibinfo{person}{Nicolas Heess}, {and} \bibinfo{person}{Josh Merel}.} \bibinfo{year}{2020}\natexlab{}.
\newblock \showarticletitle{{C}o{M}ic: Complementary Task Learning \& Mimicry for Reusable Skills}. In \bibinfo{booktitle}{\emph{Proceedings of the 37th International Conference on Machine Learning}} \emph{(\bibinfo{series}{Proceedings of Machine Learning Research}, Vol.~\bibinfo{volume}{119})}, \bibfield{editor}{\bibinfo{person}{Hal~Daumé III} {and} \bibinfo{person}{Aarti Singh}} (Eds.). \bibinfo{publisher}{PMLR}, \bibinfo{pages}{4105--4115}.
\newblock
\urldef\tempurl%
\url{https://proceedings.mlr.press/v119/hasenclever20a.html}
\showURL{%
\tempurl}


\bibitem[Hayes et~al\mbox{.}(2022)]%
        {hayes2022practical}
\bibfield{author}{\bibinfo{person}{Conor~F. Hayes}, \bibinfo{person}{Roxana R{\u{a}}dulescu}, \bibinfo{person}{Eugenio Bargiacchi}, \bibinfo{person}{Johan K{\"a}llstr{\"o}m}, \bibinfo{person}{Matthew Macfarlane}, \bibinfo{person}{Mathieu Reymond}, \bibinfo{person}{Timothy Verstraeten}, \bibinfo{person}{Luisa~M. Zintgraf}, \bibinfo{person}{Richard Dazeley}, \bibinfo{person}{Fredrik Heintz}, \bibinfo{person}{Enda Howley}, \bibinfo{person}{Athirai~A. Irissappane}, \bibinfo{person}{Patrick Mannion}, \bibinfo{person}{Ann Now{\'e}}, \bibinfo{person}{Gabriel Ramos}, \bibinfo{person}{Marcello Restelli}, \bibinfo{person}{Peter Vamplew}, {and} \bibinfo{person}{Diederik~M. Roijers}.} \bibinfo{year}{2022}\natexlab{}.
\newblock \showarticletitle{A practical guide to multi-objective reinforcement learning and planning}.
\newblock \bibinfo{journal}{\emph{Autonomous Agents and Multi-Agent Systems}} \bibinfo{volume}{36}, \bibinfo{number}{1} (\bibinfo{date}{13 Apr} \bibinfo{year}{2022}), \bibinfo{pages}{26}.
\newblock
\showISSN{1573-7454}
\href{https://doi.org/10.1007/s10458-022-09552-y}{doi:\nolinkurl{10.1007/s10458-022-09552-y}}


\bibitem[He et~al\mbox{.}(2024)]%
        {he2024hover}
\bibfield{author}{\bibinfo{person}{Tairan He}, \bibinfo{person}{Wenli Xiao}, \bibinfo{person}{Toru Lin}, \bibinfo{person}{Zhengyi Luo}, \bibinfo{person}{Zhenjia Xu}, \bibinfo{person}{Zhenyu Jiang}, \bibinfo{person}{Changliu Liu}, \bibinfo{person}{Guanya Shi}, \bibinfo{person}{Xiaolong Wang}, \bibinfo{person}{Linxi Fan}, {and} \bibinfo{person}{Yuke Zhu}.} \bibinfo{year}{2024}\natexlab{}.
\newblock \showarticletitle{HOVER: Versatile Neural Whole-Body Controller for Humanoid Robots}.
\newblock \bibinfo{journal}{\emph{arXiv preprint arXiv:2410.21229}} (\bibinfo{year}{2024}).
\newblock


\bibitem[Heess et~al\mbox{.}(2017)]%
        {heess2017emergence}
\bibfield{author}{\bibinfo{person}{Nicolas Heess}, \bibinfo{person}{Dhruva Tb}, \bibinfo{person}{Srinivasan Sriram}, \bibinfo{person}{Jay Lemmon}, \bibinfo{person}{Josh Merel}, \bibinfo{person}{Greg Wayne}, \bibinfo{person}{Yuval Tassa}, \bibinfo{person}{Tom Erez}, \bibinfo{person}{Ziyu Wang}, \bibinfo{person}{SM Eslami}, {et~al\mbox{.}}} \bibinfo{year}{2017}\natexlab{}.
\newblock \showarticletitle{Emergence of locomotion behaviours in rich environments}.
\newblock \bibinfo{journal}{\emph{arXiv preprint arXiv:1707.02286}} (\bibinfo{year}{2017}).
\newblock


\bibitem[Heess et~al\mbox{.}(2015)]%
        {heess2015continuous}
\bibfield{author}{\bibinfo{person}{Nicolas Heess}, \bibinfo{person}{Gregory Wayne}, \bibinfo{person}{David Silver}, \bibinfo{person}{Timothy Lillicrap}, \bibinfo{person}{Tom Erez}, {and} \bibinfo{person}{Yuval Tassa}.} \bibinfo{year}{2015}\natexlab{}.
\newblock \showarticletitle{Learning Continuous Control Policies by Stochastic Value Gradients}. In \bibinfo{booktitle}{\emph{Advances in Neural Information Processing Systems}}, \bibfield{editor}{\bibinfo{person}{C.~Cortes}, \bibinfo{person}{N.~Lawrence}, \bibinfo{person}{D.~Lee}, \bibinfo{person}{M.~Sugiyama}, {and} \bibinfo{person}{R.~Garnett}} (Eds.), Vol.~\bibinfo{volume}{28}. \bibinfo{publisher}{Curran Associates, Inc.}
\newblock
\urldef\tempurl%
\url{https://proceedings.neurips.cc/paper_files/paper/2015/file/148510031349642de5ca0c544f31b2ef-Paper.pdf}
\showURL{%
\tempurl}


\bibitem[Ho and Ermon(2016)]%
        {ho2016generative}
\bibfield{author}{\bibinfo{person}{Jonathan Ho} {and} \bibinfo{person}{Stefano Ermon}.} \bibinfo{year}{2016}\natexlab{}.
\newblock \showarticletitle{Generative adversarial imitation learning}.
\newblock \bibinfo{journal}{\emph{Advances in neural information processing systems}}  \bibinfo{volume}{29} (\bibinfo{year}{2016}).
\newblock


\bibitem[Hodgins et~al\mbox{.}(1995)]%
        {hodgins1995animating}
\bibfield{author}{\bibinfo{person}{Jessica~K. Hodgins}, \bibinfo{person}{Wayne~L. Wooten}, \bibinfo{person}{David~C. Brogan}, {and} \bibinfo{person}{James~F. O'Brien}.} \bibinfo{year}{1995}\natexlab{}.
\newblock \showarticletitle{Animating human athletics}. In \bibinfo{booktitle}{\emph{Proceedings of the 22nd Annual Conference on Computer Graphics and Interactive Techniques}} \emph{(\bibinfo{series}{SIGGRAPH '95})}. \bibinfo{publisher}{Association for Computing Machinery}, \bibinfo{address}{New York, NY, USA}, \bibinfo{pages}{71–78}.
\newblock
\showISBNx{0897917014}
\href{https://doi.org/10.1145/218380.218414}{doi:\nolinkurl{10.1145/218380.218414}}


\bibitem[Ibarz et~al\mbox{.}(2021)]%
        {ibarz2021how}
\bibfield{author}{\bibinfo{person}{Julian Ibarz}, \bibinfo{person}{Jie Tan}, \bibinfo{person}{Chelsea Finn}, \bibinfo{person}{Mrinal Kalakrishnan}, \bibinfo{person}{Peter Pastor}, {and} \bibinfo{person}{Sergey Levine}.} \bibinfo{year}{2021}\natexlab{}.
\newblock \showarticletitle{How to train your robot with deep reinforcement learning: lessons we have learned}.
\newblock \bibinfo{journal}{\emph{The International Journal of Robotics Research}} \bibinfo{volume}{40}, \bibinfo{number}{4-5} (\bibinfo{year}{2021}), \bibinfo{pages}{698--721}.
\newblock
\href{https://doi.org/10.1177/0278364920987859}{doi:\nolinkurl{10.1177/0278364920987859}}


\bibitem[Kim et~al\mbox{.}(2024)]%
        {kim2024stagewise}
\bibfield{author}{\bibinfo{person}{Dohyeong Kim}, \bibinfo{person}{Hyeokjin Kwon}, \bibinfo{person}{Junseok Kim}, \bibinfo{person}{Gunmin Lee}, {and} \bibinfo{person}{Songhwai Oh}.} \bibinfo{year}{2024}\natexlab{}.
\newblock \bibinfo{title}{Stage-Wise Reward Shaping for Acrobatic Robots: A Constrained Multi-Objective Reinforcement Learning Approach}.
\newblock
\showeprint[arxiv]{2409.15755}~[cs.RO]
\urldef\tempurl%
\url{https://arxiv.org/abs/2409.15755}
\showURL{%
\tempurl}


\bibitem[Lee et~al\mbox{.}(2010)]%
        {lee2010data}
\bibfield{author}{\bibinfo{person}{Yoonsang Lee}, \bibinfo{person}{Sungeun Kim}, {and} \bibinfo{person}{Jehee Lee}.} \bibinfo{year}{2010}\natexlab{}.
\newblock \showarticletitle{Data-driven biped control}. In \bibinfo{booktitle}{\emph{ACM SIGGRAPH 2010 Papers}} (Los Angeles, California) \emph{(\bibinfo{series}{SIGGRAPH '10})}. \bibinfo{publisher}{Association for Computing Machinery}, \bibinfo{address}{New York, NY, USA}, Article \bibinfo{articleno}{129}, \bibinfo{numpages}{8}~pages.
\newblock
\showISBNx{9781450302104}
\href{https://doi.org/10.1145/1833349.1781155}{doi:\nolinkurl{10.1145/1833349.1781155}}


\bibitem[Liu et~al\mbox{.}(2015)]%
        {liu2015sampling}
\bibfield{author}{\bibinfo{person}{Libin Liu}, \bibinfo{person}{KangKang Yin}, {and} \bibinfo{person}{Baining Guo}.} \bibinfo{year}{2015}\natexlab{}.
\newblock \showarticletitle{Improving Sampling-based Motion Control}.
\newblock \bibinfo{journal}{\emph{Computer Graphics Forum}} \bibinfo{volume}{34}, \bibinfo{number}{2} (\bibinfo{year}{2015}), \bibinfo{pages}{415--423}.
\newblock
\href{https://doi.org/10.1111/cgf.12571}{doi:\nolinkurl{10.1111/cgf.12571}}
\showeprint{https://onlinelibrary.wiley.com/doi/pdf/10.1111/cgf.12571}


\bibitem[Liu et~al\mbox{.}(2012)]%
        {liu2012terrain}
\bibfield{author}{\bibinfo{person}{Libin Liu}, \bibinfo{person}{KangKang Yin}, \bibinfo{person}{Michiel van~de Panne}, {and} \bibinfo{person}{Baining Guo}.} \bibinfo{year}{2012}\natexlab{}.
\newblock \showarticletitle{Terrain runner: control, parameterization, composition, and planning for highly dynamic motions}.
\newblock \bibinfo{journal}{\emph{ACM Transactions on Graphics (TOG)}} \bibinfo{volume}{31}, \bibinfo{number}{6} (\bibinfo{year}{2012}), \bibinfo{pages}{154}.
\newblock


\bibitem[Liu et~al\mbox{.}(2010)]%
        {liu2010samcon}
\bibfield{author}{\bibinfo{person}{Libin Liu}, \bibinfo{person}{KangKang Yin}, \bibinfo{person}{Michiel van~de Panne}, \bibinfo{person}{Tianjia Shao}, {and} \bibinfo{person}{Weiwei Xu}.} \bibinfo{year}{2010}\natexlab{}.
\newblock \showarticletitle{Sampling-based contact-rich motion control}.
\newblock \bibinfo{journal}{\emph{ACM Trans. Graph.}} \bibinfo{volume}{29}, \bibinfo{number}{4}, Article \bibinfo{articleno}{128} (\bibinfo{date}{July} \bibinfo{year}{2010}), \bibinfo{numpages}{10}~pages.
\newblock
\showISSN{0730-0301}
\href{https://doi.org/10.1145/1778765.1778865}{doi:\nolinkurl{10.1145/1778765.1778865}}


\bibitem[Luo et~al\mbox{.}(2023)]%
        {luo2023phc}
\bibfield{author}{\bibinfo{person}{Zhengyi Luo}, \bibinfo{person}{Jinkun Cao}, \bibinfo{person}{Alexander~W. Winkler}, \bibinfo{person}{Kris Kitani}, {and} \bibinfo{person}{Weipeng Xu}.} \bibinfo{year}{2023}\natexlab{}.
\newblock \showarticletitle{Perpetual Humanoid Control for Real-time Simulated Avatars}. In \bibinfo{booktitle}{\emph{International Conference on Computer Vision (ICCV)}}.
\newblock


\bibitem[Makoviychuk et~al\mbox{.}(2021)]%
        {makoviychuk2021isaacgym}
\bibfield{author}{\bibinfo{person}{Viktor Makoviychuk}, \bibinfo{person}{Lukasz Wawrzyniak}, \bibinfo{person}{Yunrong Guo}, \bibinfo{person}{Michelle Lu}, \bibinfo{person}{Kier Storey}, \bibinfo{person}{Miles Macklin}, \bibinfo{person}{David Hoeller}, \bibinfo{person}{Nikita Rudin}, \bibinfo{person}{Arthur Allshire}, \bibinfo{person}{Ankur Handa}, {and} \bibinfo{person}{Gavriel State}.} \bibinfo{year}{2021}\natexlab{}.
\newblock \showarticletitle{Isaac Gym: High Performance {GPU} Based Physics Simulation For Robot Learning}. In \bibinfo{booktitle}{\emph{Proceedings of The Thirty-fifth Conference on Neural Information Processing Systems Datasets and Benchmarks Track}}.
\newblock
\urldef\tempurl%
\url{https://openreview.net/forum?id=fgFBtYgJQX_}
\showURL{%
\tempurl}


\bibitem[Merel et~al\mbox{.}(2019)]%
        {merel2018neural}
\bibfield{author}{\bibinfo{person}{Josh Merel}, \bibinfo{person}{Leonard Hasenclever}, \bibinfo{person}{Alexandre Galashov}, \bibinfo{person}{Arun Ahuja}, \bibinfo{person}{Vu Pham}, \bibinfo{person}{Greg Wayne}, \bibinfo{person}{Yee~Whye Teh}, {and} \bibinfo{person}{Nicolas Heess}.} \bibinfo{year}{2019}\natexlab{}.
\newblock \showarticletitle{Neural Probabilistic Motor Primitives for Humanoid Control}. In \bibinfo{booktitle}{\emph{International Conference on Learning Representations}}.
\newblock
\urldef\tempurl%
\url{https://openreview.net/forum?id=BJl6TjRcY7}
\showURL{%
\tempurl}


\bibitem[Merel et~al\mbox{.}(2017)]%
        {merel2017learning}
\bibfield{author}{\bibinfo{person}{Josh Merel}, \bibinfo{person}{Yuval Tassa}, \bibinfo{person}{Dhruva TB}, \bibinfo{person}{Sriram Srinivasan}, \bibinfo{person}{Jay Lemmon}, \bibinfo{person}{Ziyu Wang}, \bibinfo{person}{Greg Wayne}, {and} \bibinfo{person}{Nicolas Heess}.} \bibinfo{year}{2017}\natexlab{}.
\newblock \bibinfo{title}{Learning human behaviors from motion capture by adversarial imitation}.
\newblock
\showeprint[arxiv]{1707.02201}~[cs.RO]
\urldef\tempurl%
\url{https://arxiv.org/abs/1707.02201}
\showURL{%
\tempurl}


\bibitem[Merel et~al\mbox{.}(2020)]%
        {merel2020catch}
\bibfield{author}{\bibinfo{person}{Josh Merel}, \bibinfo{person}{Saran Tunyasuvunakool}, \bibinfo{person}{Arun Ahuja}, \bibinfo{person}{Yuval Tassa}, \bibinfo{person}{Leonard Hasenclever}, \bibinfo{person}{Vu Pham}, \bibinfo{person}{Tom Erez}, \bibinfo{person}{Greg Wayne}, {and} \bibinfo{person}{Nicolas Heess}.} \bibinfo{year}{2020}\natexlab{}.
\newblock \showarticletitle{Catch \& Carry: reusable neural controllers for vision-guided whole-body tasks}.
\newblock \bibinfo{journal}{\emph{ACM Trans. Graph.}} \bibinfo{volume}{39}, \bibinfo{number}{4}, Article \bibinfo{articleno}{39} (\bibinfo{date}{Aug.} \bibinfo{year}{2020}), \bibinfo{numpages}{14}~pages.
\newblock
\showISSN{0730-0301}
\href{https://doi.org/10.1145/3386569.3392474}{doi:\nolinkurl{10.1145/3386569.3392474}}


\bibitem[Mescheder et~al\mbox{.}(2018)]%
        {mescheder2018training}
\bibfield{author}{\bibinfo{person}{Lars Mescheder}, \bibinfo{person}{Sebastian Nowozin}, {and} \bibinfo{person}{Andreas Geiger}.} \bibinfo{year}{2018}\natexlab{}.
\newblock \showarticletitle{Which Training Methods for GANs do actually Converge?}. In \bibinfo{booktitle}{\emph{Proceedings of the 35th International Conference on Machine Learning}}.
\newblock


\bibitem[Mnih et~al\mbox{.}(2016)]%
        {mnih2016async}
\bibfield{author}{\bibinfo{person}{Volodymyr Mnih}, \bibinfo{person}{Adria~Puigdomenech Badia}, \bibinfo{person}{Mehdi Mirza}, \bibinfo{person}{Alex Graves}, \bibinfo{person}{Timothy Lillicrap}, \bibinfo{person}{Tim Harley}, \bibinfo{person}{David Silver}, {and} \bibinfo{person}{Koray Kavukcuoglu}.} \bibinfo{year}{2016}\natexlab{}.
\newblock \showarticletitle{Asynchronous Methods for Deep Reinforcement Learning}. In \bibinfo{booktitle}{\emph{Proceedings of The 33rd International Conference on Machine Learning}} \emph{(\bibinfo{series}{Proceedings of Machine Learning Research}, Vol.~\bibinfo{volume}{48})}, \bibfield{editor}{\bibinfo{person}{Maria~Florina Balcan} {and} \bibinfo{person}{Kilian~Q. Weinberger}} (Eds.). \bibinfo{publisher}{PMLR}, \bibinfo{address}{New York, New York, USA}, \bibinfo{pages}{1928--1937}.
\newblock
\urldef\tempurl%
\url{https://proceedings.mlr.press/v48/mniha16.html}
\showURL{%
\tempurl}


\bibitem[Mnih et~al\mbox{.}(2015)]%
        {mnih2015human}
\bibfield{author}{\bibinfo{person}{Volodymyr Mnih}, \bibinfo{person}{Koray Kavukcuoglu}, \bibinfo{person}{David Silver}, \bibinfo{person}{Andrei~A Rusu}, \bibinfo{person}{Joel Veness}, \bibinfo{person}{Marc~G Bellemare}, \bibinfo{person}{Alex Graves}, \bibinfo{person}{Martin Riedmiller}, \bibinfo{person}{Andreas~K Fidjeland}, \bibinfo{person}{Georg Ostrovski}, {et~al\mbox{.}}} \bibinfo{year}{2015}\natexlab{}.
\newblock \showarticletitle{Human-level control through deep reinforcement learning}.
\newblock \bibinfo{journal}{\emph{nature}} \bibinfo{volume}{518}, \bibinfo{number}{7540} (\bibinfo{year}{2015}), \bibinfo{pages}{529--533}.
\newblock


\bibitem[Mordatch et~al\mbox{.}(2010)]%
        {mordatch2010robust}
\bibfield{author}{\bibinfo{person}{Igor Mordatch}, \bibinfo{person}{Martin de Lasa}, {and} \bibinfo{person}{Aaron Hertzmann}.} \bibinfo{year}{2010}\natexlab{}.
\newblock \showarticletitle{Robust physics-based locomotion using low-dimensional planning}. In \bibinfo{booktitle}{\emph{ACM SIGGRAPH 2010 Papers}} (Los Angeles, California) \emph{(\bibinfo{series}{SIGGRAPH '10})}. \bibinfo{publisher}{Association for Computing Machinery}, \bibinfo{address}{New York, NY, USA}, Article \bibinfo{articleno}{71}, \bibinfo{numpages}{8}~pages.
\newblock
\showISBNx{9781450302104}
\href{https://doi.org/10.1145/1833349.1778808}{doi:\nolinkurl{10.1145/1833349.1778808}}


\bibitem[Mordatch et~al\mbox{.}(2012)]%
        {mordatch2012disovery}
\bibfield{author}{\bibinfo{person}{Igor Mordatch}, \bibinfo{person}{Emanuel Todorov}, {and} \bibinfo{person}{Zoran Popovi\'{c}}.} \bibinfo{year}{2012}\natexlab{}.
\newblock \showarticletitle{Discovery of complex behaviors through contact-invariant optimization}.
\newblock \bibinfo{journal}{\emph{ACM Trans. Graph.}} \bibinfo{volume}{31}, \bibinfo{number}{4}, Article \bibinfo{articleno}{43} (\bibinfo{date}{July} \bibinfo{year}{2012}), \bibinfo{numpages}{8}~pages.
\newblock
\showISSN{0730-0301}
\href{https://doi.org/10.1145/2185520.2185539}{doi:\nolinkurl{10.1145/2185520.2185539}}


\bibitem[Nowozin et~al\mbox{.}(2016)]%
        {nowozin2016fgan}
\bibfield{author}{\bibinfo{person}{Sebastian Nowozin}, \bibinfo{person}{Botond Cseke}, {and} \bibinfo{person}{Ryota Tomioka}.} \bibinfo{year}{2016}\natexlab{}.
\newblock \showarticletitle{f-GAN: training generative neural samplers using variational divergence minimization}. In \bibinfo{booktitle}{\emph{Proceedings of the 30th International Conference on Neural Information Processing Systems}} (Barcelona, Spain) \emph{(\bibinfo{series}{NIPS'16})}. \bibinfo{publisher}{Curran Associates Inc.}, \bibinfo{address}{Red Hook, NY, USA}, \bibinfo{pages}{271–279}.
\newblock
\showISBNx{9781510838819}


\bibitem[Park et~al\mbox{.}(2019)]%
        {park2019learning}
\bibfield{author}{\bibinfo{person}{Soohwan Park}, \bibinfo{person}{Hoseok Ryu}, \bibinfo{person}{Seyoung Lee}, \bibinfo{person}{Sunmin Lee}, {and} \bibinfo{person}{Jehee Lee}.} \bibinfo{year}{2019}\natexlab{}.
\newblock \showarticletitle{Learning predict-and-simulate policies from unorganized human motion data}.
\newblock \bibinfo{journal}{\emph{ACM Trans. Graph.}} \bibinfo{volume}{38}, \bibinfo{number}{6}, Article \bibinfo{articleno}{205} (\bibinfo{date}{Nov.} \bibinfo{year}{2019}), \bibinfo{numpages}{11}~pages.
\newblock
\showISSN{0730-0301}
\href{https://doi.org/10.1145/3355089.3356501}{doi:\nolinkurl{10.1145/3355089.3356501}}


\bibitem[Peng et~al\mbox{.}(2018a)]%
        {peng2018deepmimic}
\bibfield{author}{\bibinfo{person}{Xue~Bin Peng}, \bibinfo{person}{Pieter Abbeel}, \bibinfo{person}{Sergey Levine}, {and} \bibinfo{person}{Michiel van~de Panne}.} \bibinfo{year}{2018}\natexlab{a}.
\newblock \showarticletitle{DeepMimic: example-guided deep reinforcement learning of physics-based character skills}.
\newblock \bibinfo{journal}{\emph{ACM Trans. Graph.}} \bibinfo{volume}{37}, \bibinfo{number}{4}, Article \bibinfo{articleno}{143} (\bibinfo{date}{July} \bibinfo{year}{2018}), \bibinfo{numpages}{14}~pages.
\newblock
\showISSN{0730-0301}
\href{https://doi.org/10.1145/3197517.3201311}{doi:\nolinkurl{10.1145/3197517.3201311}}


\bibitem[Peng et~al\mbox{.}(2017)]%
        {peng2017deeploco}
\bibfield{author}{\bibinfo{person}{Xue~Bin Peng}, \bibinfo{person}{Glen Berseth}, \bibinfo{person}{Kangkang Yin}, {and} \bibinfo{person}{Michiel Van De~Panne}.} \bibinfo{year}{2017}\natexlab{}.
\newblock \showarticletitle{DeepLoco: dynamic locomotion skills using hierarchical deep reinforcement learning}.
\newblock \bibinfo{journal}{\emph{ACM Trans. Graph.}} \bibinfo{volume}{36}, \bibinfo{number}{4}, Article \bibinfo{articleno}{41} (\bibinfo{date}{July} \bibinfo{year}{2017}), \bibinfo{numpages}{13}~pages.
\newblock
\showISSN{0730-0301}
\href{https://doi.org/10.1145/3072959.3073602}{doi:\nolinkurl{10.1145/3072959.3073602}}


\bibitem[Peng et~al\mbox{.}(2019)]%
        {peng2019mcp}
\bibfield{author}{\bibinfo{person}{Xue~Bin Peng}, \bibinfo{person}{Michael Chang}, \bibinfo{person}{Grace Zhang}, \bibinfo{person}{Pieter Abbeel}, {and} \bibinfo{person}{Sergey Levine}.} \bibinfo{year}{2019}\natexlab{}.
\newblock \bibinfo{booktitle}{\emph{MCP: learning composable hierarchical control with multiplicative compositional policies}}.
\newblock \bibinfo{publisher}{Curran Associates Inc.}, \bibinfo{address}{Red Hook, NY, USA}.
\newblock


\bibitem[Peng et~al\mbox{.}(2018b)]%
        {peng2018sfv}
\bibfield{author}{\bibinfo{person}{Xue~Bin Peng}, \bibinfo{person}{Angjoo Kanazawa}, \bibinfo{person}{Jitendra Malik}, \bibinfo{person}{Pieter Abbeel}, {and} \bibinfo{person}{Sergey Levine}.} \bibinfo{year}{2018}\natexlab{b}.
\newblock \showarticletitle{SFV: reinforcement learning of physical skills from videos}.
\newblock \bibinfo{journal}{\emph{ACM Trans. Graph.}} \bibinfo{volume}{37}, \bibinfo{number}{6}, Article \bibinfo{articleno}{178} (\bibinfo{date}{Dec.} \bibinfo{year}{2018}), \bibinfo{numpages}{14}~pages.
\newblock
\showISSN{0730-0301}
\href{https://doi.org/10.1145/3272127.3275014}{doi:\nolinkurl{10.1145/3272127.3275014}}


\bibitem[Peng et~al\mbox{.}(2021)]%
        {peng2021amp}
\bibfield{author}{\bibinfo{person}{Xue~Bin Peng}, \bibinfo{person}{Ze Ma}, \bibinfo{person}{Pieter Abbeel}, \bibinfo{person}{Sergey Levine}, {and} \bibinfo{person}{Angjoo Kanazawa}.} \bibinfo{year}{2021}\natexlab{}.
\newblock \showarticletitle{AMP: adversarial motion priors for stylized physics-based character control}.
\newblock \bibinfo{journal}{\emph{ACM Trans. Graph.}} \bibinfo{volume}{40}, \bibinfo{number}{4}, Article \bibinfo{articleno}{144} (\bibinfo{date}{jul} \bibinfo{year}{2021}), \bibinfo{numpages}{20}~pages.
\newblock
\showISSN{0730-0301}
\href{https://doi.org/10.1145/3450626.3459670}{doi:\nolinkurl{10.1145/3450626.3459670}}


\bibitem[Reallusion(2023)]%
        {reallusion}
\bibfield{author}{\bibinfo{person}{Reallusion}.} \bibinfo{year}{2023}\natexlab{}.
\newblock \bibinfo{booktitle}{\emph{3D Animation and 2D Cartoons Made Simple.}}
\newblock
\urldef\tempurl%
\url{https://www.reallusion.com}
\showURL{%
\tempurl}
\newblock
\shownote{https://actorcore.reallusion.com/motion/pack/studio-mocap-sword-and-shield-stunts, https://actorcore.reallusion.com/motion/pack/studio-mocap-sword-and-shield-moves, https://actorcore.reallusion.com/3d-motion/pack/studio-mocap-hero-motion, https://actorcore.reallusion.com/3d-motion/pack/studio-mocap-girl-dance, https://actorcore.reallusion.com/3d-motion/pack/studio-mocap-evolution-of-dance-vol-1, https://actorcore.reallusion.com/3d-motion/pack/studio-mocap-evolution-of-dance-vol-2, https://actorcore.reallusion.com/3d-motion/pack/iclone-motion-pack---street-dance-locking}.


\bibitem[Roijers et~al\mbox{.}(2013)]%
        {roijers2013survey}
\bibfield{author}{\bibinfo{person}{Diederik~M. Roijers}, \bibinfo{person}{Peter Vamplew}, \bibinfo{person}{Shimon Whiteson}, {and} \bibinfo{person}{Richard Dazeley}.} \bibinfo{year}{2013}\natexlab{}.
\newblock \showarticletitle{A Survey of Multi-Objective Sequential Decision-Making}.
\newblock \bibinfo{journal}{\emph{J. Artificial Intelligence Research}} \bibinfo{volume}{48}, \bibinfo{number}{1} (\bibinfo{date}{Oct.} \bibinfo{year}{2013}), \bibinfo{pages}{67--113}.
\newblock
\showISSN{1076-9757}


\bibitem[Schulman et~al\mbox{.}(2015a)]%
        {schulman2015tppo}
\bibfield{author}{\bibinfo{person}{John Schulman}, \bibinfo{person}{Sergey Levine}, \bibinfo{person}{Pieter Abbeel}, \bibinfo{person}{Michael Jordan}, {and} \bibinfo{person}{Philipp Moritz}.} \bibinfo{year}{2015}\natexlab{a}.
\newblock \showarticletitle{Trust Region Policy Optimization}. In \bibinfo{booktitle}{\emph{Proceedings of the 32nd International Conference on Machine Learning}} \emph{(\bibinfo{series}{Proceedings of Machine Learning Research}, Vol.~\bibinfo{volume}{37})}, \bibfield{editor}{\bibinfo{person}{Francis Bach} {and} \bibinfo{person}{David Blei}} (Eds.). \bibinfo{publisher}{PMLR}, \bibinfo{address}{Lille, France}, \bibinfo{pages}{1889--1897}.
\newblock
\urldef\tempurl%
\url{https://proceedings.mlr.press/v37/schulman15.html}
\showURL{%
\tempurl}


\bibitem[Schulman et~al\mbox{.}(2015b)]%
        {schulman2015high}
\bibfield{author}{\bibinfo{person}{John Schulman}, \bibinfo{person}{Philipp Moritz}, \bibinfo{person}{Sergey Levine}, \bibinfo{person}{Michael Jordan}, {and} \bibinfo{person}{Pieter Abbeel}.} \bibinfo{year}{2015}\natexlab{b}.
\newblock \showarticletitle{High-dimensional continuous control using generalized advantage estimation}.
\newblock \bibinfo{journal}{\emph{arXiv preprint arXiv:1506.02438}} (\bibinfo{year}{2015}).
\newblock


\bibitem[Schulman et~al\mbox{.}(2017)]%
        {schulman2017ppo}
\bibfield{author}{\bibinfo{person}{John Schulman}, \bibinfo{person}{Filip Wolski}, \bibinfo{person}{Prafulla Dhariwal}, \bibinfo{person}{Alec Radford}, {and} \bibinfo{person}{Oleg Klimov}.} \bibinfo{year}{2017}\natexlab{}.
\newblock \showarticletitle{Proximal Policy Optimization Algorithms}.
\newblock \bibinfo{journal}{\emph{CoRR}}  \bibinfo{volume}{abs/1707.06347} (\bibinfo{year}{2017}).
\newblock
\showeprint[arXiv]{1707.06347}
\urldef\tempurl%
\url{http://arxiv.org/abs/1707.06347}
\showURL{%
\tempurl}


\bibitem[Schulz et~al\mbox{.}(2018)]%
        {schulz2018interactive}
\bibfield{author}{\bibinfo{person}{Adriana Schulz}, \bibinfo{person}{Harrison Wang}, \bibinfo{person}{Eitan Grinspun}, \bibinfo{person}{Justin Solomon}, {and} \bibinfo{person}{Wojciech Matusik}.} \bibinfo{year}{2018}\natexlab{}.
\newblock \showarticletitle{Interactive exploration of design trade-offs}.
\newblock \bibinfo{journal}{\emph{ACM Trans. Graph.}} \bibinfo{volume}{37}, \bibinfo{number}{4}, Article \bibinfo{articleno}{131} (\bibinfo{date}{July} \bibinfo{year}{2018}), \bibinfo{numpages}{14}~pages.
\newblock
\showISSN{0730-0301}
\href{https://doi.org/10.1145/3197517.3201385}{doi:\nolinkurl{10.1145/3197517.3201385}}


\bibitem[Serifi et~al\mbox{.}(2024)]%
        {serifi2024vmp}
\bibfield{author}{\bibinfo{person}{Agon Serifi}, \bibinfo{person}{Ruben Grandia}, \bibinfo{person}{Espen Knoop}, \bibinfo{person}{Markus Gross}, {and} \bibinfo{person}{Moritz B\"{a}cher}.} \bibinfo{year}{2024}\natexlab{}.
\newblock \showarticletitle{VMP: Versatile Motion Priors for Robustly Tracking Motion on Physical Characters}. In \bibinfo{booktitle}{\emph{Proceedings of the ACM SIGGRAPH/Eurographics Symposium on Computer Animation}} (Montreal, Quebec, Canada) \emph{(\bibinfo{series}{SCA '24})}. \bibinfo{publisher}{Eurographics Association}, \bibinfo{address}{Goslar, DEU}, \bibinfo{pages}{1–11}.
\newblock
\href{https://doi.org/10.1111/cgf.15175}{doi:\nolinkurl{10.1111/cgf.15175}}


\bibitem[Sharon and van~de Panne(2005)]%
        {sharon2005synthesis}
\bibfield{author}{\bibinfo{person}{D. Sharon} {and} \bibinfo{person}{M. van~de Panne}.} \bibinfo{year}{2005}\natexlab{}.
\newblock \showarticletitle{Synthesis of Controllers for Stylized Planar Bipedal Walking}. In \bibinfo{booktitle}{\emph{Proceedings of the 2005 IEEE International Conference on Robotics and Automation}}. \bibinfo{pages}{2387--2392}.
\newblock
\href{https://doi.org/10.1109/ROBOT.2005.1570470}{doi:\nolinkurl{10.1109/ROBOT.2005.1570470}}


\bibitem[Sok et~al\mbox{.}(2007)]%
        {sok2007simulating}
\bibfield{author}{\bibinfo{person}{Kwang~Won Sok}, \bibinfo{person}{Manmyung Kim}, {and} \bibinfo{person}{Jehee Lee}.} \bibinfo{year}{2007}\natexlab{}.
\newblock \showarticletitle{Simulating biped behaviors from human motion data}. In \bibinfo{booktitle}{\emph{ACM SIGGRAPH 2007 Papers}} (San Diego, California) \emph{(\bibinfo{series}{SIGGRAPH '07})}. \bibinfo{publisher}{Association for Computing Machinery}, \bibinfo{address}{New York, NY, USA}, \bibinfo{pages}{107–es}.
\newblock
\showISBNx{9781450378369}
\href{https://doi.org/10.1145/1275808.1276511}{doi:\nolinkurl{10.1145/1275808.1276511}}


\bibitem[Sutton and Barto(2018)]%
        {sutton2018rl}
\bibfield{author}{\bibinfo{person}{Richard~S. Sutton} {and} \bibinfo{person}{Andrew~G. Barto}.} \bibinfo{year}{2018}\natexlab{}.
\newblock \bibinfo{booktitle}{\emph{Reinforcement learning: An introduction} (\bibinfo{edition}{second} ed.)}.
\newblock \bibinfo{publisher}{The MIT Press}, \bibinfo{address}{Cambridge, MA, USA}.
\newblock


\bibitem[Tang et~al\mbox{.}(2024)]%
        {tang2024humanmimic}
\bibfield{author}{\bibinfo{person}{Annan Tang}, \bibinfo{person}{Takuma Hiraoka}, \bibinfo{person}{Naoki Hiraoka}, \bibinfo{person}{Fan Shi}, \bibinfo{person}{Kento Kawaharazuka}, \bibinfo{person}{Kunio Kojima}, \bibinfo{person}{Kei Okada}, {and} \bibinfo{person}{Masayuki Inaba}.} \bibinfo{year}{2024}\natexlab{}.
\newblock \showarticletitle{Humanmimic: Learning natural locomotion and transitions for humanoid robot via wasserstein adversarial imitation}. In \bibinfo{booktitle}{\emph{2024 IEEE International Conference on Robotics and Automation (ICRA)}}. IEEE, \bibinfo{pages}{13107--13114}.
\newblock


\bibitem[Tessler et~al\mbox{.}(2024)]%
        {tessler2024maskedmimic}
\bibfield{author}{\bibinfo{person}{Chen Tessler}, \bibinfo{person}{Yunrong Guo}, \bibinfo{person}{Ofir Nabati}, \bibinfo{person}{Gal Chechik}, {and} \bibinfo{person}{Xue~Bin Peng}.} \bibinfo{year}{2024}\natexlab{}.
\newblock \showarticletitle{MaskedMimic: Unified Physics-Based Character Control Through Masked Motion Inpainting}.
\newblock \bibinfo{journal}{\emph{ACM Trans. Graph.}} \bibinfo{volume}{43}, \bibinfo{number}{6}, Article \bibinfo{articleno}{209} (\bibinfo{date}{Nov.} \bibinfo{year}{2024}), \bibinfo{numpages}{21}~pages.
\newblock
\showISSN{0730-0301}
\href{https://doi.org/10.1145/3687951}{doi:\nolinkurl{10.1145/3687951}}


\bibitem[Tessler et~al\mbox{.}(2023)]%
        {tessler2023calm}
\bibfield{author}{\bibinfo{person}{Chen Tessler}, \bibinfo{person}{Yoni Kasten}, \bibinfo{person}{Yunrong Guo}, \bibinfo{person}{Shie Mannor}, \bibinfo{person}{Gal Chechik}, {and} \bibinfo{person}{Xue~Bin Peng}.} \bibinfo{year}{2023}\natexlab{}.
\newblock \showarticletitle{CALM: Conditional Adversarial Latent Models for Directable Virtual Characters}. In \bibinfo{booktitle}{\emph{ACM SIGGRAPH 2023 Conference Proceedings}} (Los Angeles, CA, USA) \emph{(\bibinfo{series}{SIGGRAPH '23})}. \bibinfo{publisher}{Association for Computing Machinery}, \bibinfo{address}{New York, NY, USA}.
\newblock
\showISBNx{9798400701597}
\href{https://doi.org/10.1145/3588432.3591541}{doi:\nolinkurl{10.1145/3588432.3591541}}


\bibitem[Torabi et~al\mbox{.}(2018a)]%
        {torabi2018behavioral}
\bibfield{author}{\bibinfo{person}{Faraz Torabi}, \bibinfo{person}{Garrett Warnell}, {and} \bibinfo{person}{Peter Stone}.} \bibinfo{year}{2018}\natexlab{a}.
\newblock \showarticletitle{Behavioral cloning from observation}. In \bibinfo{booktitle}{\emph{Proceedings of the 27th International Joint Conference on Artificial Intelligence}} (Stockholm, Sweden) \emph{(\bibinfo{series}{IJCAI'18})}. \bibinfo{publisher}{AAAI Press}, \bibinfo{pages}{4950–4957}.
\newblock
\showISBNx{9780999241127}


\bibitem[Torabi et~al\mbox{.}(2018b)]%
        {torabi2018generative}
\bibfield{author}{\bibinfo{person}{Faraz Torabi}, \bibinfo{person}{Garrett Warnell}, {and} \bibinfo{person}{Peter Stone}.} \bibinfo{year}{2018}\natexlab{b}.
\newblock \showarticletitle{Generative adversarial imitation from observation}.
\newblock \bibinfo{journal}{\emph{arXiv preprint arXiv:1807.06158}} (\bibinfo{year}{2018}).
\newblock


\bibitem[Van~Moffaert and Now\'e(2014)]%
        {vanmoffaert&nowe2014}
\bibfield{author}{\bibinfo{person}{Kristof Van~Moffaert} {and} \bibinfo{person}{Ann Now\'e}.} \bibinfo{year}{2014}\natexlab{}.
\newblock \showarticletitle{Multi-Objective Reinforcement Learning Using Sets of {P}areto Dominating Policies}.
\newblock \bibinfo{journal}{\emph{Journal of Machine Learning Research}} \bibinfo{volume}{15}, \bibinfo{number}{1} (\bibinfo{year}{2014}), \bibinfo{pages}{3483--3512}.
\newblock
\showISSN{1532-4435}


\bibitem[Wang et~al\mbox{.}(2020)]%
        {wang2020unicon}
\bibfield{author}{\bibinfo{person}{Tingwu Wang}, \bibinfo{person}{Yunrong Guo}, \bibinfo{person}{Maria Shugrina}, {and} \bibinfo{person}{Sanja Fidler}.} \bibinfo{year}{2020}\natexlab{}.
\newblock \bibinfo{title}{UniCon: Universal Neural Controller For Physics-based Character Motion}.
\newblock
\showeprint[arxiv]{2011.15119}~[cs.GR]


\bibitem[Won et~al\mbox{.}(2020)]%
        {won2020scadiver}
\bibfield{author}{\bibinfo{person}{Jungdam Won}, \bibinfo{person}{Deepak Gopinath}, {and} \bibinfo{person}{Jessica Hodgins}.} \bibinfo{year}{2020}\natexlab{}.
\newblock \showarticletitle{A Scalable Approach to Control Diverse Behaviors for Physically Simulated Characters}.
\newblock \bibinfo{journal}{\emph{ACM Trans. Graph.}} \bibinfo{volume}{39}, \bibinfo{number}{4}, Article \bibinfo{articleno}{33} (\bibinfo{year}{2020}).
\newblock
\urldef\tempurl%
\url{https://doi.org/10.1145/3386569.3392381}
\showURL{%
\tempurl}


\bibitem[Won et~al\mbox{.}(2022)]%
        {won2022physics}
\bibfield{author}{\bibinfo{person}{Jungdam Won}, \bibinfo{person}{Deepak Gopinath}, {and} \bibinfo{person}{Jessica Hodgins}.} \bibinfo{year}{2022}\natexlab{}.
\newblock \showarticletitle{Physics-based character controllers using conditional VAEs}.
\newblock \bibinfo{journal}{\emph{ACM Trans. Graph.}} \bibinfo{volume}{41}, \bibinfo{number}{4}, Article \bibinfo{articleno}{96} (\bibinfo{date}{July} \bibinfo{year}{2022}), \bibinfo{numpages}{12}~pages.
\newblock
\showISSN{0730-0301}
\href{https://doi.org/10.1145/3528223.3530067}{doi:\nolinkurl{10.1145/3528223.3530067}}


\bibitem[Won et~al\mbox{.}(2017)]%
        {won2017dragon}
\bibfield{author}{\bibinfo{person}{Jungdam Won}, \bibinfo{person}{Jongho Park}, \bibinfo{person}{Kwanyu Kim}, {and} \bibinfo{person}{Jehee Lee}.} \bibinfo{year}{2017}\natexlab{}.
\newblock \showarticletitle{How to train your dragon: example-guided control of flapping flight}.
\newblock \bibinfo{journal}{\emph{ACM Trans. Graph.}} \bibinfo{volume}{36}, \bibinfo{number}{6}, Article \bibinfo{articleno}{198} (\bibinfo{date}{Nov.} \bibinfo{year}{2017}), \bibinfo{numpages}{13}~pages.
\newblock
\showISSN{0730-0301}
\href{https://doi.org/10.1145/3130800.3130833}{doi:\nolinkurl{10.1145/3130800.3130833}}


\bibitem[Wu et~al\mbox{.}(2022)]%
        {wu2022daydreamerworldmodelsphysical}
\bibfield{author}{\bibinfo{person}{Philipp Wu}, \bibinfo{person}{Alejandro Escontrela}, \bibinfo{person}{Danijar Hafner}, \bibinfo{person}{Ken Goldberg}, {and} \bibinfo{person}{Pieter Abbeel}.} \bibinfo{year}{2022}\natexlab{}.
\newblock \bibinfo{title}{DayDreamer: World Models for Physical Robot Learning}.
\newblock
\showeprint[arxiv]{2206.14176}~[cs.RO]
\urldef\tempurl%
\url{https://arxiv.org/abs/2206.14176}
\showURL{%
\tempurl}


\bibitem[Xiang and Li(2020)]%
        {xiang2020revisiting}
\bibfield{author}{\bibinfo{person}{Sitao Xiang} {and} \bibinfo{person}{Hao Li}.} \bibinfo{year}{2020}\natexlab{}.
\newblock \showarticletitle{Revisiting the continuity of rotation representations in neural networks}.
\newblock \bibinfo{journal}{\emph{arXiv preprint arXiv:2006.06234}} (\bibinfo{year}{2020}).
\newblock


\bibitem[Xu et~al\mbox{.}(2020)]%
        {xu2020prediction}
\bibfield{author}{\bibinfo{person}{Jie Xu}, \bibinfo{person}{Yunsheng Tian}, \bibinfo{person}{Pingchuan Ma}, \bibinfo{person}{Daniela Rus}, \bibinfo{person}{Shinjiro Sueda}, {and} \bibinfo{person}{Wojciech Matusik}.} \bibinfo{year}{2020}\natexlab{}.
\newblock \showarticletitle{Prediction-Guided Multi-Objective Reinforcement Learning for Continuous Robot Control}. In \bibinfo{booktitle}{\emph{Proceedings of the 37th International Conference on Machine Learning (ICML)}}.
\newblock


\bibitem[Xu and Karamouzas(2021)]%
        {xu2021gan}
\bibfield{author}{\bibinfo{person}{Pei Xu} {and} \bibinfo{person}{Ioannis Karamouzas}.} \bibinfo{year}{2021}\natexlab{}.
\newblock \showarticletitle{A GAN-Like Approach for Physics-Based Imitation Learning and Interactive Character Control}.
\newblock \bibinfo{journal}{\emph{Proc. ACM Comput. Graph. Interact. Tech.}} \bibinfo{volume}{4}, \bibinfo{number}{3}, Article \bibinfo{articleno}{44} (\bibinfo{date}{Sept.} \bibinfo{year}{2021}), \bibinfo{numpages}{22}~pages.
\newblock
\href{https://doi.org/10.1145/3480148}{doi:\nolinkurl{10.1145/3480148}}


\bibitem[Xu et~al\mbox{.}(2023)]%
        {xu2023composite}
\bibfield{author}{\bibinfo{person}{Pei Xu}, \bibinfo{person}{Xiumin Shang}, \bibinfo{person}{Victor Zordan}, {and} \bibinfo{person}{Ioannis Karamouzas}.} \bibinfo{year}{2023}\natexlab{}.
\newblock \showarticletitle{Composite Motion Learning with Task Control}.
\newblock \bibinfo{journal}{\emph{ACM Trans. Graph.}} \bibinfo{volume}{42}, \bibinfo{number}{4}, Article \bibinfo{articleno}{93} (\bibinfo{date}{July} \bibinfo{year}{2023}), \bibinfo{numpages}{16}~pages.
\newblock
\showISSN{0730-0301}
\href{https://doi.org/10.1145/3592447}{doi:\nolinkurl{10.1145/3592447}}


\bibitem[Yang et~al\mbox{.}(2019)]%
        {yang2019generalized}
\bibfield{author}{\bibinfo{person}{Runzhe Yang}, \bibinfo{person}{Xingyuan Sun}, {and} \bibinfo{person}{Karthik Narasimhan}.} \bibinfo{year}{2019}\natexlab{}.
\newblock \bibinfo{booktitle}{\emph{A generalized algorithm for multi-objective reinforcement learning and policy adaptation}}.
\newblock \bibinfo{publisher}{Curran Associates Inc.}, \bibinfo{address}{Red Hook, NY, USA}.
\newblock


\bibitem[Yao et~al\mbox{.}(2022)]%
        {yao2022controlvae}
\bibfield{author}{\bibinfo{person}{Heyuan Yao}, \bibinfo{person}{Zhenhua Song}, \bibinfo{person}{Baoquan Chen}, {and} \bibinfo{person}{Libin Liu}.} \bibinfo{year}{2022}\natexlab{}.
\newblock \showarticletitle{ControlVAE: Model-Based Learning of Generative Controllers for Physics-Based Characters}.
\newblock \bibinfo{journal}{\emph{ACM Trans. Graph.}} \bibinfo{volume}{41}, \bibinfo{number}{6}, Article \bibinfo{articleno}{183} (\bibinfo{date}{Nov.} \bibinfo{year}{2022}), \bibinfo{numpages}{16}~pages.
\newblock
\showISSN{0730-0301}
\href{https://doi.org/10.1145/3550454.3555434}{doi:\nolinkurl{10.1145/3550454.3555434}}


\bibitem[Yu et~al\mbox{.}(2018)]%
        {yu2018energy}
\bibfield{author}{\bibinfo{person}{Wenhao Yu}, \bibinfo{person}{Greg Turk}, {and} \bibinfo{person}{C.~Karen Liu}.} \bibinfo{year}{2018}\natexlab{}.
\newblock \showarticletitle{Learning symmetric and low-energy locomotion}.
\newblock \bibinfo{journal}{\emph{ACM Trans. Graph.}} \bibinfo{volume}{37}, \bibinfo{number}{4}, Article \bibinfo{articleno}{144} (\bibinfo{date}{July} \bibinfo{year}{2018}), \bibinfo{numpages}{12}~pages.
\newblock
\showISSN{0730-0301}
\href{https://doi.org/10.1145/3197517.3201397}{doi:\nolinkurl{10.1145/3197517.3201397}}


\bibitem[Zhou et~al\mbox{.}(2019)]%
        {zhou2019continuity}
\bibfield{author}{\bibinfo{person}{Yi Zhou}, \bibinfo{person}{Connelly Barnes}, \bibinfo{person}{Jingwan Lu}, \bibinfo{person}{Jimei Yang}, {and} \bibinfo{person}{Hao Li}.} \bibinfo{year}{2019}\natexlab{}.
\newblock \showarticletitle{On the continuity of rotation representations in neural networks}. In \bibinfo{booktitle}{\emph{Proceedings of the IEEE/CVF Conference on Computer Vision and Pattern Recognition}}. \bibinfo{pages}{5745--5753}.
\newblock


\bibitem[Zhu et~al\mbox{.}(2023)]%
        {zhu2023neural}
\bibfield{author}{\bibinfo{person}{Qingxu Zhu}, \bibinfo{person}{He Zhang}, \bibinfo{person}{Mengting Lan}, {and} \bibinfo{person}{Lei Han}.} \bibinfo{year}{2023}\natexlab{}.
\newblock \showarticletitle{Neural Categorical Priors for Physics-Based Character Control}.
\newblock \bibinfo{journal}{\emph{ACM Trans. Graph.}} \bibinfo{volume}{42}, \bibinfo{number}{6}, Article \bibinfo{articleno}{178} (\bibinfo{date}{Dec.} \bibinfo{year}{2023}), \bibinfo{numpages}{16}~pages.
\newblock
\showISSN{0730-0301}
\href{https://doi.org/10.1145/3618397}{doi:\nolinkurl{10.1145/3618397}}


\end{thebibliography}


\end{document}